\documentclass{article}

\usepackage{microtype}
\usepackage{graphicx}
\usepackage{subcaption}
\usepackage{booktabs} %
\usepackage{multirow}

\usepackage{hyperref}

\usepackage[accepted]{icml2023}

\usepackage{amsmath}
\usepackage{amssymb}
\usepackage{mathtools}
\usepackage{amsthm}
\usepackage{physics}
\usepackage{nicefrac}
\usepackage[algo2e,inoutnumbered,algoruled,vlined]{algorithm2e} %
\usepackage{tikz}
\usetikzlibrary{fit,calc}
\usepackage{comment}
\newcommand*{\tikzmk}[1]{\tikz[remember picture,overlay,] \node (#1) {};\ignorespaces}
\newcommand{\boxit}[1]{\tikz[remember picture,overlay]{\node[yshift=3pt,fill=#1,opacity=.25,fit={(A)($(B)+(.95\linewidth,.8\baselineskip)$)}] {};}\ignorespaces}

\usepackage[capitalize,noabbrev]{cleveref}

\theoremstyle{plain}

\theoremstyle{definition}

\theoremstyle{remark}

\usepackage[textsize=tiny]{todonotes}

\newcommand{\id}{\mathrm{id}}
\newcommand{\F}{\mathcal{F}}
\newcommand{\R}{\mathbb{R}}
\newcommand{\Rd}{\mathbb{R}^d}
\newcommand{\Sd}{\mathbb{S}^{d-1}}
\newcommand{\thetapush}{\theta^*_{\sharp}}
\newcommand{\PR}{\mathcal{P}(\mathbb{R})}
\newcommand{\PRd}{\mathcal{P}(\mathbb{R}^d)}
\newcommand{\PtRd}{\mathcal{P}_2(\mathbb{R}^d)}

\newcommand{\PtX}{\mathcal{P}_2(\mathcal{X})}

\newcommand{\PtXY}{\mathcal{P}_2(\mathcal{X}\times \mathcal{Y})}

\newcommand{\vectheta}{\begin{bmatrix}\theta_x\\\theta_y\end{bmatrix}}

\newcommand{\data}{\mathcal{D}}
\newcommand{\simiid}{\stackrel{\mathclap{\normalfont\mbox{\tiny i.i.d.}}}{\sim}}

\DeclarePairedDelimiter\floor{\lfloor}{\rfloor}

\newcommand{\eqn}[1]{Eq.~(\ref{eqn:#1})}
\newcommand{\secref}[1]{Sec.~\ref{sec:#1}}
\newcommand{\figref}[1]{Fig.~\ref{fig:#1}}
\newcommand{\tabref}[1]{Table~\ref{table:#1}}

\newcommand{\possessivecite}[1]{\citeauthor{#1}'s \citeyearpar{#1}}

\newcommand{\method}{$\ell$-CSWF}
\newcommand{\ucmethod}{$\ell$-SWF}

\graphicspath{{./figures/}}

\hypersetup{
	colorlinks=true,
	urlcolor=magenta,
}

\begin{document}

\twocolumn[
\icmltitle{Nonparametric Generative Modeling with Conditional Sliced-Wasserstein Flows}

\begin{icmlauthorlist}
\icmlauthor{Chao Du}{sail}
\icmlauthor{Tianbo Li}{sail}
\icmlauthor{Tianyu Pang}{sail}
\icmlauthor{Shuicheng Yan}{sail}
\icmlauthor{Min Lin}{sail}
\end{icmlauthorlist}

\icmlaffiliation{sail}{Sea AI Lab, Singapore}

\icmlcorrespondingauthor{Chao Du}{duchao@sea.com}
\icmlcorrespondingauthor{Min Lin}{linmin@sea.com}

\icmlkeywords{Machine Learning, ICML}

\vskip 0.3in
]

\printAffiliationsAndNotice{}  %

\begin{abstract}

Sliced-Wasserstein Flow (SWF) is a promising approach to nonparametric generative modeling but has not been widely adopted due to its suboptimal generative quality and lack of conditional modeling capabilities. In this work, we make two major contributions to bridging this gap. First, based on a pleasant observation that (under certain conditions) the SWF of joint distributions coincides with those of conditional distributions, we propose Conditional Sliced-Wasserstein Flow (CSWF), a simple yet effective extension of SWF that enables nonparametric conditional modeling. Second, we introduce appropriate inductive biases of images into SWF with two techniques inspired by local connectivity and multiscale representation in vision research, which greatly improve the efficiency and quality of modeling images. With all the improvements, we achieve generative performance comparable with many deep parametric generative models on both conditional and unconditional tasks in a purely nonparametric fashion, demonstrating its great potential.
\end{abstract}

\section{Introduction}

Deep generative models have made several breakthroughs in recent years~\citep{brock2018large,durkan2019neural,child2020very,song2020score}, thanks to the powerful function approximation capability of deep neural networks and the various families of models tailored for probabilistic modeling. One recurring pattern in 
deep generative models
for continuous variables is to map a simple prior distribution to the data distribution (or vice versa).
Examples include
those performing single-step mappings via a neural network, such as GANs~\citep{goodfellow2014generative}, VAEs~\citep{kingma2013auto}, and normalizing flows~\citep{rezende2015variational}, and those taking iterative steps to transform the distributions characterized by ODEs~\citep{grathwohl2018ffjord} and diffusion SDEs~\citep{song2020score}. 
The latter
have achieved great success recently
due to the decomposition of complex mappings into multi-step easier ones.

While most existing models train parametric neural networks as the mapping functions, a less popular but promising alternative is to perform nonparametric mappings, for which preliminary attempts~\cite{liutkus2019sliced,dai2021sliced} have been made recently.
These works further decompose the mapping between multidimensional distributions into mappings between one-dimensional distributions, which have closed-form solutions based on the optimal transport theory~\citep{villani2008optimal}.
The closed-form solutions enable a nonparametric way to construct the mappings, which makes no (or weak) assumptions about the underlying distributions and is therefore more flexible.
A promising work is the Sliced-Wasserstein Flows (SWF)~\citep{liutkus2019sliced}, which achieves generative modeling by building nonparametric mappings to solve the gradient flows in the space of probability distribution.

Despite the potential advantages, nonparametric methods like SWF are still in the early stage of development. 
First, it is underexplored how they can be applied to various tasks besides unconditional generation, such as conditional generation and image inpainting~\cite{meng2021sdedit}. In contrast, parametric probability models have mature techniques for constructing conditional distributions and end-to-end gradient descent training, making tasks such as text-to-image generation and image inpainting more straightforward to implement~\citep{mirza2014conditional,van2016conditional,perez2018film}.
Similar mechanisms have yet to be developed for SWF, obstructing its applications in conditional modeling. Second, the quality and efficiency are still quite limited when applied to high-dimensional data such as images. While parametric models can leverage the sophisticated inductive biases of neural networks (e.g., the U-Net~\cite{ronneberger2015u} in diffusion models) and are able to process full-resolution images directly, the nonparametric counterpart either operates on low-dimensional features processed from a pre-trained auto-encoder \citep{kolouri2018sliced,liutkus2019sliced} or relies on a carefully designed patch-based approach~\citep{dai2021sliced}. Although several variants have been proposed to incorporate convolutional architectures~\citep{nguyen2022revisiting,laparra2022orthonormal}, to our knowledge, none has demonstrated empirical success in image generation.

In this work, we propose two improvements for SWF to address the above limitations.
First, we extend the framework of SWF for conditional probabilistic modeling based on an empirical observation that the collection of SWFs w.r.t.\ conditional distributions (approximately) coincide with the SWF w.r.t.\ the joint distribution, subject to a few conditions that can be easily met.
Based on this finding, we propose a simple yet effective algorithm, named \emph{Conditional Sliced-Wasserstein Flows} (CSWF), 
where we obtain conditional samples by simulating the SWF w.r.t.\ the joint distribution.
CSWF enjoys the same nonparametric advantages as SWF while being able to perform various types of conditional inference, such as class-conditional generation and image inpainting.
Second, we introduce the \emph{locally-connected projections} and \emph{pyramidal schedules} techniques to enhance the quality of image generation, motivated by the common notions of local connectivity and pyramidal representation in computer vision.
By solving optimal transport problems in domain-specific rather than isotropic directions,
we successfully incorporate visual inductive biases into SWF (and our CSWF) for image tasks.

Our method and techniques have made several remarkable achievements.
First, to the best of our knowledge, the proposed CSWF is the first nonparametric generative approach that is able to handle general conditional distribution modeling tasks.
Second, our proposed techniques greatly improve the generation quality of images and also reveals a general way to introduce inductive biases into SWF and CSWF\@.
Last but not least, we achieve comparable performance to many parametric models on both unconditional and conditional image generation tasks,
showing great promise.

\section{Related Work} \label{sec:related}

\textbf{Parametric Generative Models} Generative models are one of the core research areas in machine learning. Several classes of parametric generative models are widely studied, including GANs~\cite{goodfellow2014generative}, VAEs~\cite{kingma2013auto}, flow-based models~\citep{durkan2019neural}, autoregressive models~\cite{van2016conditional}, energy-based models~\citep{du2019implicit} and diffusion/score-based models~\cite{ho2020denoising,song2020score}.
In this work, we instead study a completely different approach to generative modeling using nonparametric methods.

\textbf{Generative Models based on Optimal Transport} Optimal transport (OT) have been widely adopted in generative modeling~\cite{arjovsky2017wasserstein,gulrajani2017improved,tolstikhin2017wasserstein,kolouri2018sliced,deshpande2018generative,deshpande2019max,meng2019large,wu2019sliced,knop2020cramer,nguyen2020distributional,nguyen2020improving,nguyen2022revisiting,bonet2021sliced,nguyen2022amortized}.
\citet{meng2019large} build iterative normalizing flows where they identify the most informative directions to construct OT maps in each layer. \citet{arjovsky2017wasserstein,wu2019sliced} propose to train GANs using different distances based on OT.
\citet{nguyen2020distributional,nguyen2022hierarchical} propose difference variants of the sliced-Wasserstein distance and apply them on generative models.
These works adopt OT in different dimensions, while they are all parametric and based on end-to-end training.

\textbf{Nonparametric Generative Models}
A less popular area of research is nonparametric generative models, which holds a lot of potential.
These methods utilize tools from nonparametric statistics such as kernel methods~\citep{shi2018spectral,li2017gradient,zhou2020nonparametric}, Gaussianization~\cite{NIPS2000_3c947bc2}, and independent component analysis~\cite{laparra2011iterative}.
\citet{meng2020gaussianization} propose a trainable Gaussianization layer via kernel density estimation and random rotations.
SINF~\cite{dai2021sliced} iteratively transform between Gaussian and data distribution using sliced OT\@.
Note, however, that none of these nonparametric generative models can perform conditional inference.

\textbf{Conditional Generative Models}
Generating samples conditioned on additional input information is crucial for achieving controllable generation.
Prior research has successfully demonstrated class-conditional image generation~\cite{mirza2014conditional,sohn2015learning}.
Recent advancements in this area have yielded notable success in synthesizing high-quality images from textual input~\cite{ramesh2022hierarchical,saharia2022photorealistic,rombach2022high,hertz2022prompt,feng2022ernie,yu2022scaling}.
The conditional generative capabilities in these models are facilitated by well-established pipelines that construct parametric conditional probability distributions using deep neural networks.
In contrast, our work achieves conditional generation within a nonparametric framework, which is significantly different.

\section{Preliminaries} \label{sec:back}

We briefly review some prior knowledge, including optimal transport, the Wasserstein space, and gradient flows.
Most results presented here hold under certain mild assumptions that can be easily met in practical problems and we have omitted them for simplicity.
For more details,
we refer the readers to \citet{ambrosio2005gradient,villani2008optimal}.

\textbf{Notations} We denote by $\mathcal{P}(\mathcal{X})$ the set of probability distributions supported on $\mathcal{X}\subseteq\Rd$.
Given $p\in\mathcal{P}(\mathcal{X}_1)$ and a measurable function $T:\mathcal{X}_1\rightarrow\mathcal{X}_2$, we denote by $q=T_{\sharp}p\in\mathcal{P}(\mathcal{X}_2)$ the pushforward distribution, defined as $q(E)=p(T^{-1}(E))$ for all Borel set $E$ of $\mathcal{X}_2$.
We slightly abuse the notation and denote both the probability distribution and its probability density function (if exists) by $p$.

\subsection{Optimal Transport} \label{sec:back:OT}

Given two probability distributions $p,q\in\PRd$
and a cost function $c:\Rd\times\Rd\rightarrow[0,\infty]$,
the optimal transport (OT) theory~\cite{monge1781memoire,kantorovich2006translocation} studies the problem of finding a distribution $\gamma\in\Gamma(p,q)$ such that $\int c(x,x')\dd\gamma(x,x')$ is minimal,
where $\Gamma(p,q)$ is the set of all \emph{transport plans} between $p$ and $q$, i.e., the set of all probability distributions on $\Rd\times\Rd$ with marginals $p$ and $q$.

Under mild conditions, 
the solution to the OT problem exists and is strongly connected with its dual formulation: $\min_{\gamma\in\Gamma (p,q)}\int c \dd\gamma = \max_{\psi\in L^1 (p)}\left\{\int \psi\dd p + \int \psi^c\dd q \right\}$,\footnote{$L^1 (p)$ is the set of absolutely integrable functions under $p$ and $\psi^c(x')\triangleq\inf_{x}c(x,x')-\psi(x)$ is called the $c$-transform of $\psi$.}
where a solution $\psi$ to the RHS is called a \emph{Kantorovich potential} between $p$ and $q$.
In particular, for the quadratic cost $c(x,x')=\|x-x'\|_2^2$,
the above equation is realized (under certain conditions) by a unique optimal transport plan $\gamma^*$ and a unique (up to an additive constant) Kantorovich potential $\psi$,
and they are related through $\gamma^*=(\id, T)_{\sharp}p$, where $T(x)=x-\nabla\psi(x)$~\cite{benamou2000computational}.
The function $T:\Rd\rightarrow\Rd$ is called the optimal \emph{transport map} as it depicts how to transport $p$ onto $q$, i.e., $q=T_{\sharp}p$.
While the optimal transport map $T$
is generally intractable,
in one-dimensional cases, i.e., $p,q\in\PR$, it has a closed-form of $T=F_{q}^{-1}\circ F_{p}$,
where $F_{p}$ and $F_{q}$ are the cumulative distribution functions (CDF) of $p$ and $q$, respectively.

\subsection{Wasserstein Distance and Wasserstein Space} \label{sec:back:wasserstein}

For the cost $c(x,x')=\|x-x'\|_2^2$,
the OT problem naturally defines a distance, called the $2$-Wasserstein distance:
\begin{equation}\label{eqn:WD}
W_2 (p,q)\triangleq\left(\min_{\gamma\in\Gamma (p,q)}\int\|x-x'\|_2^2\dd\gamma(x,x')\right)^{\nicefrac{1}{2}}.
\end{equation}
For $W_2 (p,q)$ to be finite, it is convenient to consider $\PtRd\triangleq\left\{p\in\PRd:\int\|x\|_2^2\dd p(x)<\infty\right\}$, which is the subset of all probability distributions on $\Rd$ with finite second moments.
The set $\PtRd$ equipped with the $2$-Wasserstein distance $W_2$ forms an important metric space for probability distributions,
called the \emph{Wasserstein space}.

\subsection{Gradient Flows in the Wasserstein Space} \label{sec:back:GF}

Gradient flows in metric spaces are analogous to the steepest descent curves in the classical Euclidean space.
Given a functional $\F:\PtRd\rightarrow\R$,
a gradient flow of $\F$ in the Wasserstein space is an absolutely continuous curve $(p_t)_{t\geq0}$ that minimizes $\F$ as fast as possible~\cite{santambrogio2017euclidean}.

The Wasserstein gradient flows are shown to be strongly connected with partial differential equations (PDE)~\cite{jordan1998variational}.
In particular,
it is shown that (under proper conditions)
the Wasserstein gradient flows $(p_t)_t$ coincide with the solutions of the \emph{continuity equation} $\frac{\partial}{\partial t}p_t + \nabla\cdot(p_tv_t)=0$, where 
$v_t:\Rd\rightarrow\Rd$ is a time-dependent velocity field~\cite{ambrosio2005gradient}.
Moreover, the solution of the continuity equation can be represented by $p_t=(X_t)_\sharp p_0, \forall t\geq0$, where $X_t(x)$ is defined by the characteristic system of ODEs $\frac{\dd}{\dd t}X_t(x)=v_t(X_t(x))$, $X_0(x)=x, \forall x\in\Rd$~\cite{bers1964partial}.
The PDE formulation and the characteristic system of ODEs provide us with an important perspective to analyze and simulate the Wasserstein gradient flows.

\subsection{Sliced-Wasserstein Flows} \label{sec:back:SWF}

The tractability of the one-dimensional OT and Wasserstein distance motivates the definition of the sliced-Wasserstein distance~\cite{rabin2011wasserstein}.
For any $\theta\in\Sd$ (the unit sphere in $\Rd$), we denote by $\theta^*:\Rd\rightarrow\R$ the orthogonal projection, defined as $\theta^*(x)\triangleq\theta^\top x$.
Then, counting all the Wasserstein distance between the projected distributions $\thetapush p,\thetapush q\in\PR$
leads to the sliced-Wasserstein distance:
\begin{equation}\label{eqn:SWD}
SW_2 (p,q)\triangleq\left(\int_{\Sd}W_2^2(\thetapush p,\thetapush q)\dd\lambda(\theta)\right)^{\nicefrac{1}{2}},
\end{equation}
where $\lambda(\theta)$ is the uniform distribution on the sphere $\Sd$. The $SW_2$ has many similar properties as the $W_2$~\cite{bonnotte2013unidimensional}
and can be estimated with Monte Carlo methods. It is therefore often used as an alternative to the $W_2$ in practical problems~\cite{kolouri2018sliced,deshpande2018generative}.

In this paper, we will focus on the Wasserstein gradient flows of functionals $\F(\cdot)=\frac{1}{2}SW^2_2(\cdot,q)$,
where $q$ is a target distribution. 
\citet{bonnotte2013unidimensional} 
proves that (under regularity conditions on $p_0$ and $q$) such Wasserstein gradient flows $ (p_t)_{t\geq0}$ satisfy the continuity equation (in a weak sense):
\begin{equation}\label{eqn:swfpde}
\begin{aligned}
    & ~~~\frac{\partial p_t(x)}{\partial t} + \nabla\cdot(p_t(x)v_t(x))=0, \\ 
    & v_t(x)\triangleq-\int_{\Sd}\psi'_{t,\theta}(\theta^\top x)\cdot\theta~\dd\lambda(\theta),
\end{aligned}
\end{equation}
where $\psi_{t,\theta}$ denotes the Kantorovich potential between the (one-dimensional) projected distributions $\thetapush p_t$ and $\thetapush q$. Moreover, according to \secref{back:OT} the optimal transport map from $\thetapush p_t$ to $\thetapush q$ is given by $T_{t,\theta}=F_{\thetapush q}^{-1}\circ F_{\thetapush p_t}$, which gives
$\psi'_{t,\theta}(z) = z - T_{t,\theta}(z) = z - F_{\thetapush q}^{-1}\circ F_{\thetapush p_t}(z)$.

\citet{liutkus2019sliced} refers to it as the \emph{sliced-Wasserstein flow} (SWF) and adapts it into a practical algorithm for building generative models of an unknown target distribution $q$ (assuming access to i.i.d.\ samples).
By simulating a similar PDE to \eqn{swfpde},\footnote{The authors originally consider the entropy-regularized SWFs, leading to a similar PDE, with an extra Laplacian term that is often ignored when modeling real data. See more details in Appendix~\ref{app:swf}.}
it transforms a bunch of particles sampled from $p_0$ (e.g., a Gaussian distribution) to match the target distribution $q$.
We recap more details in Appendix~\ref{app:swf}.

\section{Conditional Sliced-Wasserstein Flows}\label{sec:cswf}

In this section, we present an extended framework of SWFs for \emph{conditional} probability distributions and accordingly propose a practical nonparametric method for \emph{conditional} generative modeling.

Formally, given a dataset $\mathcal{D}\triangleq\{(x_i,y_i)\}_{i=1}^N$ representing $N$ i.i.d.\ samples from the target distribution $q\in\PtXY$, where $\mathcal{X}\subseteq\Rd$ and $\mathcal{Y}\subseteq\R^{l}$ are two related domains (e.g., $\mathcal{X}$ the image space and $\mathcal{Y}$ the set of labels), we aim to model the conditional distributions $q_y\triangleq q(\cdot|y)\in\PtX$ for all $y\in\mathcal{Y}$.
We assume the marginal distribution $ q(y)$ is known.

\subsection{Conditional Sliced-Wasserstein Flows}\label{sec:cswf:naivecswf}

A straightforward idea is to consider a SWF in $\PtX$ with the target distribution $q_y$ for each $y\in\mathcal{Y}$ separately, which we denote by $(p_{y,t})_{t\geq0}$ and refer to as the \emph{conditional SWF given} $y$.
Then with a suitable initial $p_{y,0}\!\!\in\!\mathcal{P}_2(\!\mathcal{X})$, it satisfies:
\begin{equation}\label{eqn:cswfpde}
    \begin{aligned}
    & ~~~\frac{\partial p_{y,t}(x)}{\partial t} + \nabla\cdot(p_{y,t}(x)v_{y,t}(x))=0, \\ 
    & v_{y,t}(x)\triangleq-\int_{\mathbb{S}^{d-1}}\psi'_{y,t,\theta}\left( \theta^\top x\right)\cdot\theta~\dd\lambda(\theta),
\end{aligned}
\end{equation}
where $\psi_{y,t,\theta}$ denotes the Kantorovich potential between the projected conditional distributions $\thetapush p_{y,t}$ and $\thetapush q_y$.
Samples from $q_y$ can be drawn if we can simulate the PDE (\ref{eqn:cswfpde}).

However, modeling the conditional SWFs for all $y\in\mathcal{Y}$ separately with \possessivecite{liutkus2019sliced} algorithm can be impractical for at least two reasons.
First, it is only feasible for the cases where $\mathcal{Y}$ is a finite set, and every $y$ appears in $\data$ often enough since a different split of the dataset $\mathcal{D}_y\triangleq\mathcal{D}\cap(\mathcal{X}\times\{y\})$ is required for each $y$.
Second, even for a finite $\mathcal{Y}$, the knowledge in different $\mathcal{D}_y$ cannot be shared, making it inefficient and unscalable when $|\mathcal{Y}|$ is large or the distribution over $\mathcal{Y}$ is highly imbalanced.

To overcome these difficulties, it is crucial to enable knowledge sharing and generalization abilities among $\mathcal{Y}$ by exploiting the global information from the joint distribution $q$,
rather than solely the conditional information from $q_y$.

\subsection{Conditional SWFs via the Joint SWF}\label{sec:cswf:jcswf}

We instead consider the SWF $(p_t)_{t\geq0}$ in $\PtXY$ with the target being the joint distribution $q$.
We refer to it as the \emph{joint SWF}, and write its corresponding PDE below:
\begin{equation}\label{eqn:jswfpde}
    \begin{aligned}
    & ~~\quad\quad\frac{\partial p_{t}(x,y)}{\partial t} + \nabla\cdot(p_{t}(x,y)v_{t}(x,y))=0, \\ 
    & v_t(x,y)=-\!\int_{\mathbb{S}^{d+l-1}}\!\psi'_{t,\theta}\left(\theta_x^\top x+\theta_y^\top y\right)\cdot\vectheta\dd\lambda(\theta),
\end{aligned}
\end{equation}
where $\theta=[\theta_x^\top,\theta_y^\top]^\top\in\mathbb{S}^{d+l-1}$ is $(d+l)$-dimensional, $\theta_x\in\Rd$, $\theta_y\in\R^l$,
and $\psi_{t,\theta}$ is the Kantorovich potential between $\thetapush p_t$ and $\thetapush q$.
Note that here $v_t:\R^{d+l}\rightarrow \R^{d+l}$ is a vector field on $\mathcal{X}\times\mathcal{Y}$.
We denote the $\mathcal{X}$- and $\mathcal{Y}$-components of $v_t(x,y)$ by $v^{\mathcal{X}}_t(x,y)$ and $v^{\mathcal{Y}}_t(x,y)$, respectively.

At first glance, the joint SWF $(p_t)_{t\geq0}$ may only provide us with a possibility to sample from $q$, but is not obviously helpful for modeling conditional distributions.
Interestingly, our empirical observation shows that under the assumption that (i)
$p_0(y)=q(y)$ and (ii) the target conditional distribution $q_y$ changes slowly enough w.r.t.\ $y$,
then for all $t\geq0$, we have $v^{\mathcal{X}}_t(x,y)\approx v_{y,t}(x)$ and $v^{\mathcal{Y}}_t(x,y)\approx 0$.
We are unable to provide a rigorous theoretical justification for the time being. We instead include an illustration in Appendix~\ref{app:illustration}.

Intuitively, this means that if the assumptions are met,
the evolution of distributions $(p_t)_{t\geq0}$ characterized by the joint SWF can be factorized into two levels. 
First, the marginal distributions $p_t(y)$ remain unchanged for all $t\geq0$ since the velocity has zero $\mathcal{Y}$-component.
Second, for each $y$, the evolution of the conditional distributions $(p_t(x|y))_{t\geq0}$ coincides with the evolution of $(p_{y,t})_{t\geq0}$ in the conditional SWF given $y$ described in \secref{cswf:naivecswf}.

Moreover, if we simulate the continuity equation~(\ref{eqn:jswfpde}) of the joint SWF with particle-based methods (e.g., using the characteristic system), then $v^{\mathcal{Y}}_t(x,y)\approx 0$ implies that the $\mathcal{Y}$-component of each particle will almost stand still, and $v^{\mathcal{X}}_t(x,y)\approx v_{y,t}(x)$ implies that the $\mathcal{X}$-component of each particle will 
move just as if we are simulating the conditional SWF given $y$.
This provides us with an elegant way to practically model conditional distributions through the joint SWF, as described in the next section.

\subsection{Practical Algorithm}\label{sec:cswf:algo}

Based on our observation, 
we propose a practical algorithm for conditional probabilistic modeling, dubbed the conditional sliced-Wasserstein flows (CSWF).
The basic idea is to first adjust the target distribution $q$ and initialize $p_0$ properly so that the 
assumptions
are (approximately) met, and then to
simulate the joint SWF with a particle-based method.

\textbf{Initialization}
To satisfy the assumption made in \secref{cswf:jcswf},
we can always define a new target distribution $q'\triangleq L_\sharp q$ with $L(x,y)\triangleq(x,\xi y)$, that is, $q'$ is obtained using a simple change of variable that scales the $\mathcal{Y}$-component with a number $\xi>1$, which we call the \emph{amplifier}.
Intuitively, scaling the $\mathcal{Y}$-component by $\xi$ will make the change of $q(x|y)$ (w.r.t.\ $y$) $\xi$ times slower.
The conditions required can thus be approximately satisfied with a large enough $\xi$.
In practice, given a dataset $\mathcal{D}=\{(x_i,y_i)\}_{i=1}^N$, this simply means that all we need is to use $\{(x_i,\xi y_i)\}_{i=1}^N$ for subsequent processing.
We thus assume below w.l.o.g.\ that $q$ has met the condition, to keep the notations consistent.
Then, we set the initial distribution as $p_0(x,y)=q(y)p_0(x)$, where $p_0(x)=\mathcal{N}(x;0,I_d)$ is Gaussian. This ensures the initial marginal distribution over $\mathcal{Y}$ is aligned with the target $q$.

\textbf{Particle System}
The solution $(p_t)_{t\geq0}$ to the continuity equ\-ation~(\ref{eqn:jswfpde}) can be represented by $(Z_t)_\sharp p_0,\forall t\geq0$, where $Z_t\!:\R^{d+l}\rightarrow\R^{d+l}$ denotes the mapping from $(x,y)$ to $Z_t(x,y)$ characterized by the ODE $\dd Z_t(x,y)=v_t(Z_t(x,y))\dd t$ with initial condition $Z_0(x,y)\triangleq(x,y)$ (see \secref{back:GF}). This intuitively means that we can sample from $p_t$ by simulating the ODE with initial point sampled from $p_0$.
 
To estimate the velocity field $v_t$, we follow \citet{liutkus2019sliced} to consider a particle system so that $p_t$ (and thus $v_t$) can be estimated within the system.
More precisely, we consider $M$ particles $\{\bar{Z}_t^j=(\bar{X}_t^j,\bar{Y}_t^j)\in\R^{d+l}\}_{j=1}^M$, 
described by a collection of characteristic ODEs:
\begin{equation}
    \dd \bar{Z}_t^j=\hat{v}_t(\bar{Z}^j_t)\dd t, \quad j=1,\dots,M,
\end{equation}
where $\hat{v}_t$ is the velocity field estimated with the empirical distributions $\hat{p}_t\triangleq\frac{1}{M}\sum^M_{j=1}\delta_{\bar{Z}_t^j}$ and $\hat{q}\triangleq\frac{1}{N}\sum^N_{i=1}\delta_{(x_i,y_i)}$.
Specifically, given a projection $\theta$,
the projected empirical distributions $\thetapush \hat{p}_t$ and $\thetapush \hat{q}$ become two sets of scalar values.
Then, estimating $\psi'_{t,\theta}$ is essentially the problem of fitting one-dimensional distributions (i.e., $F_{\theta_{\sharp}^* \hat{q}}$ and $F_{\theta_{\sharp}^* \hat{p}_t}$).
We simply estimate the CDFs with linear interpolations between the empirical distribution functions.
Finally, the velocity is approximated using a Monte Carlo estimate of the integral:
\begin{equation}\label{eqn:v-empirical}
    \hat{v}_t(x,y)\triangleq-\frac{1}{H}\sum_{h}^H\hat{\psi}'_{t,\theta_h}\left(\theta_{h,x}^\top x+\theta_{h,y}^\top y\right)\cdot\begin{bmatrix}\theta_{h,x}\\\theta_{h,y}\end{bmatrix},
\end{equation}
where $\{\theta_h\}_{h=1}^H$ are $H$ i.i.d.\ samples from the unit sphere $\mathbb{S}^{d+l-1}$ and
$\hat{\psi}'_{t,\theta_h}$ is the derivative of the Kantorovich potential computed with the estimated CDFs:
\begin{equation}\label{eqn:kp-empirical}
    \hat{\psi}'_{t,\theta_h}(z)=z-F_{\theta_{h\sharp}^* \hat{q}}^{-1}\circ F_{\theta_{h\sharp}^* \hat{p}_t}(z).
\end{equation}

\textbf{Velocity Masking}
Although the initialization has provided a good approximation of the required conditions, the velocity may still has a small $\mathcal{Y}$-component, which can be accumulated over time $t$. We correct this by manually setting $\hat{v}^{\mathcal{Y}}_t(x,y)=0$, which means that only the $\mathcal{X}$-component of each particle is updated during the simulation.

Finally, we adopt the Euler method with step size $\eta$ to iteratively simulate the particle system (i.e., the characteristic ODEs) for $K$ steps.
The particles $\{\bar{Z}_0^j=(\bar{X}_0^j,\bar{Y}_0^j)\}_{j=1}^M$ are initialized by independently sampling $\{\bar{Y}_0^j\}_{j=1}^M$ from $q(y)$ and sampling $\{\bar{X}_0^j\}_{j=1}^M$ from $\mathcal{N}(0,I_d)$.
In cases where we do not have access to the true marginal $q(y)$, we can alternatively sample $\{\bar{Y}_0^j\}_{j=1}^M$ from the dataset (with replacement).
We describe the overall CSWF algorithm in Algorithm~\ref{algo:cswf}.
Note that by simulating the particle system, we end up with $M$ conditional samples $\{\bar{X}_K^j\}_{j=1}^M$, which we refer to as the \emph{batched samples}.
Once we have simulated a particle system, we can opt to save all the $\theta$ and the CDFs and reuse them as a model.
Then, we can generate new samples conditioned on any input $y\in\mathcal{Y}$, by following the same pipeline of Algorithm~\ref{algo:cswf} but with the \colorbox{pink!25}{shaded lines} skipped.
We refer to such samples as the \emph{offline samples}.

\begin{algorithm2e}[t]
    \SetInd{0.1ex}{1.5ex}
    \DontPrintSemicolon
    \SetKwInput{Input}{Input}
    \SetKwInput{Output}{Output}
    \Input{$\mathcal{D}=\{(x_i,y_i)\}_{i=1}^N$, $\{\bar{Y}_0^j\}_{j=1}^M$, $\xi$, $H$, $\eta$, $K$}
    \Output{$\{\bar{X}_K^j\}_{j=1}^M$}
    {\color{blue} \small \tcp{Initialize the $\mathcal{X}$-component}}
    $\{\bar{X}_0^j\}_{j=1}^M \simiid \mathcal{N}(0,I_d)$\\
    {\color{blue} \small \tcp{Discretize the ODEs}}
    \For{$k = 0,\dots, K-1$}
    {
        \tikzmk{A}\For{$h = 1,\dots,H$}
        {
        {\color{blue} \small \tcp{Generate random projections}}
        $\theta_{h} \sim \mathrm{Uniform}(\mathbb{S}^{d+l-1})$\\
        {\color{blue} \small \tcp{Estimate the CDFs}}
        $F_{\theta_{h\sharp}^* \hat{q}}=\textrm{CDF}(\{\theta_{h,x}^{~\top} x_i+\xi\cdot\theta_{h,y}^{~\top} y_i \}_{i=1}^N)$\\
        $F_{\theta_{h\sharp}^* \hat{p}_k}=\textrm{CDF}(\{\theta_{h,x}^{~\top} \bar{X}_k^j+\xi\cdot\theta_{h,y}^{~\top} \bar{Y}_0^j \}_{j=1}^M)$\\
        }\tikzmk{B}
        \boxit{pink}
        {\color{blue} \small \tcp{Update the $\mathcal{X}$-component with (\ref{eqn:v-empirical})\&(\ref{eqn:kp-empirical})}}
        $\bar{X}_{k+1}^j = \bar{X}_{k}^j - \eta \cdot\hat{v}_k^{\mathcal{X}}(\bar{X}_{k}^j, \xi\cdot Y_{0}^j)$ \hfill $j=1,\dots,M$\\
    }
    \caption{Conditional Sliced-Wasserstein Flow}
    \label{algo:cswf}
\end{algorithm2e}

\subsection{Discussions}\label{sec:cswf:discuss}

The advantages of considering the joint SWF instead of separate conditional SWFs as described in \secref{cswf:naivecswf} become more clear now.
As one can see, the generalization ability comes from interpolating the empirical CDFs. 
In our method, we always interpolate the CDFs of the projected joint distributions $\thetapush \hat{q}$,
suggesting that we are indeed generalizing across both $\mathcal{X}$ and $\mathcal{Y}$. Hence, the estimated velocity field applies to all $y\in\mathcal{Y}$ even if it does not exist in $\data$.
Moreover, the CDFs are always estimated using the entire dataset, which means the knowledge is shared for all $y\in\mathcal{Y}$.

The time complexity is discussed here.
In each step of the simulation, estimating the empirical CDFs for each projection requires sorting two sets of scalar values, with time complexity $\mathcal{O}(M\log M+N\log N)$. Therefore, the overall time complexity is $\mathcal{O}(KH(M\log M+N\log N))$ and the per-sample complexity is $\mathcal{O}(KH(\log M+\frac{N}{M}\log N))$.
For the offline samples, since querying $F_{\theta_{h\sharp}^* \hat{p}_t}$ and $F_{\theta_{h\sharp}^* \hat{q}}^{-1}$ are indeed binary search and indexing operations, the per-sample time complexity is $\mathcal{O}(KH\log M)$.
Note that the constant $KH$ can possibly be further reduced by sharing projection between steps, which is left for future work.

Similar to \citet{liutkus2019sliced}, the nonparametric nature of CSWF stems from expressing the CDFs directly with empirical data (e.g., sorted arrays of projections). This makes it fundamentally different from parametric generative models that are typically learned via (stochastic) gradient descent training and thus implies many potential advantages over them. 
Notably, when new data samples are observed, the empirical CDFs of the projected data distributions $\thetapush \hat{q}$ can be updated perfectly by only insertion operations (in $\mathcal{O}(\log N)$ time), which suggests that CSWF has great potential to be adapted to online methods and bypass the challenges associated with parametric online learning, such as catastrophic forgetting~\cite{kirkpatrick2017overcoming,french1999catastrophic}. This is also an exciting direction for follow-up research.

Finally, it is worth noting that by setting $\xi=0$ the effect of conditions is completely removed and our method falls back to an unconditional variant similar to \citet{liutkus2019sliced}.

\section{SWFs with Visual Inductive Biases}\label{sec:convcswf}

In this section, we propose to introduce appropriate inductive biases for image tasks into SWF-based methods  via \emph{locally-connected projections} and \emph{pyramidal schedules}.
The key idea is to use domain-specific projection distributions instead of the uniform distribution, thus focusing more on the OT problems in critical directions.
We shall show below how we adapt our CSWF with these techniques.
Adapting the SWF algorithm should then be straightforward.

\subsection{Locally-Connected Projections}

For an image domain $\mathcal{X}\subseteq\Rd$, We assume that $d=\mathtt{C}\times \mathtt{H}\times \mathtt{W}$, where $\mathtt{H}$ and $\mathtt{W}$ denote the height and width of the image in pixels and $\mathtt{C}$ denotes the number of channels.

\begin{figure}[t]
\centering
\includegraphics[width=\linewidth]{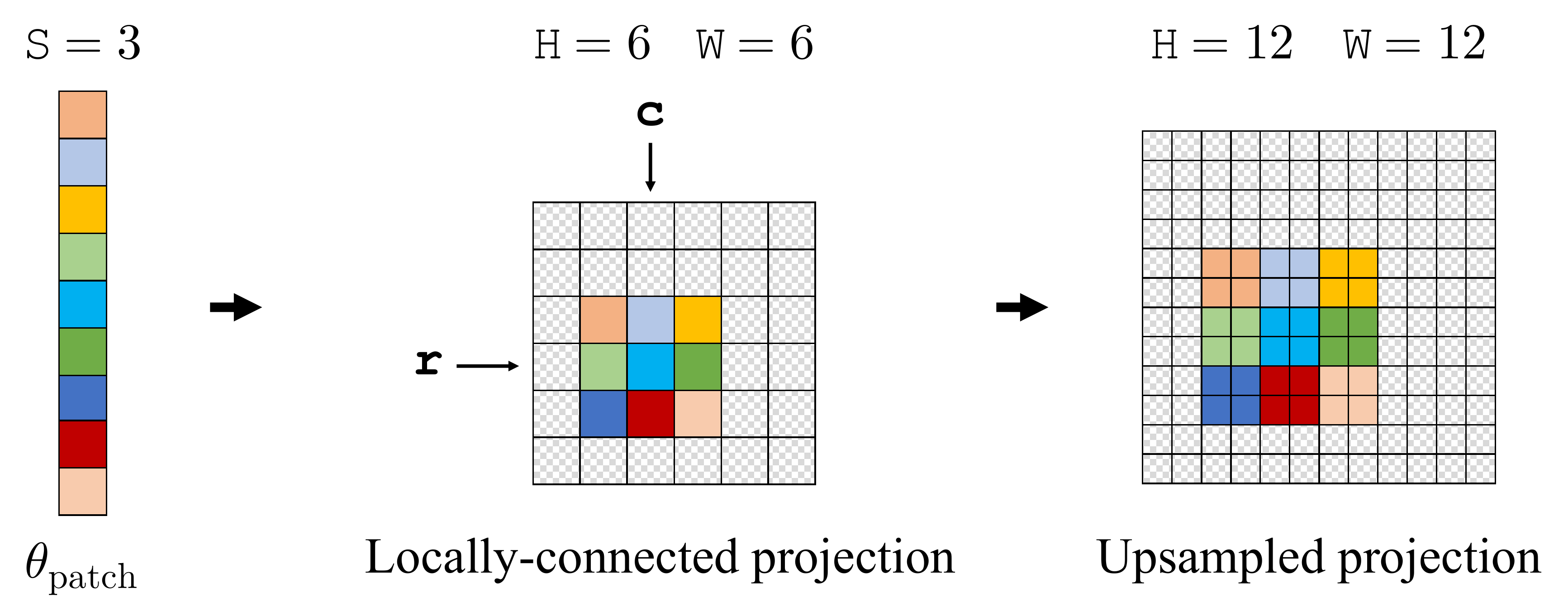}
\caption{An illustration of locally-connected projection and the upsampled projection used in pyramidal schedules.
Left panel shows an example of how we generate a projection $\theta_x$ for $6\times 6$ image with patch size $3\times 3$ and spatial position $(\mathtt{r}=4,\mathtt{c}=3)$.
Right panel shows how we upsample the projection to size $12\times12$.
}
\label{fig:conv}
\end{figure}

We observe that projecting an image $x\in\mathcal{X}$ with uniformly sampled $\theta$ is analogous to using a fully-connected layer on the flattened vector of the image in neural networks, in the sense that all pixels contribute to the projected values (or the neurons).
On the other hand, it has been widely recognized that local connectivity is one of the key features making CNNs effective for image tasks~\cite{ngiam2010tiled}.
This motivates us to use \emph{locally-connected projections}, where $\theta_x$ is made sparse so that it only projects a small patch of the image $x$.
Specifically,
given a patch size $\mathtt{S}$,
we first sample a projection $\theta_{\mathrm{patch}}$ at the patch level from the $(\mathtt{C}\times \mathtt{S}\times \mathtt{S})$-dimensional sphere.
Then, we sample a spatial position $(\mathtt{r},\mathtt{c})$ (i.e., the row and column indices), which stands for the center of the patch.
Finally, we obtain $\theta_x$ by embedding $\theta_{\mathrm{patch}}$ into a $(\mathtt{C}\times \mathtt{H}\times \mathtt{W})$-dimensional zero vector such that $\theta_x^\top x$ is 
equivalent to the projection of the $\mathtt{S}\times \mathtt{S}$ patch of $x$ centered at $(\mathtt{r},\mathtt{c})$ using $\theta_{\mathrm{patch}}$,\footnote{Note that such choices of projections result in a non-uniform distribution over $\mathbb{S}^{d-1}$ and thus do not necessarily induce a well-defined metric in $\PtRd$.
However, it can be practically effective for image data.
See more discussion in \citet{nguyen2022revisiting}.
} as illustrated in \figref{conv}.

The domain knowledge of $\mathcal{Y}$ can be incorporated into $\theta_y$ in a similar way.
To obtain the final projection $\theta$, we simply concatenate $\theta_x$ and $\theta_y$ and normalize it to ensure $\theta\in\mathbb{S}^{d+l-1}$.

\subsection{Pyramidal Schedules}

Pyramid (or multiscale) representation of images has been widely used in computer vision~\cite{adelson1984pyramid,lin2017feature}. In image classification~\cite{krizhevsky}, neural networks start from a high-resolution input with details and gradually apply subsampling to obtain low-resolution feature maps that contain high-level information. Image generation usually follows the reverse order, i.e., starts by sketching the high-level structure and completes the details gradually~\cite{ramesh2022hierarchical,jing2022subspace}. 

We thus adapt our CSWF to image tasks by introducing \emph{pyramidal schedules}, where we apply locally-connected projections at different resolutions from low to high sequentially. However, due to the dimension-preserving constraint of SWF, instead of working directly on a low-resolution image, we translate the operation to the full-sized image by upsampling the projection filter
and modifying the stride parameter accordingly. 
See \figref{conv} for an illustration and more details in \secref{exp} and Appendix~\ref{app:exp-detail}.

In the following, we refer to our CSWF combined with locally-connected projections and pyramidal schedules as the \emph{locally-connected CSWF} and denote it as \method{}.
We refer to unconditional SWF~\cite{liutkus2019sliced} combined with the same techniques as \ucmethod{}.

\section{Experiments}\label{sec:exp}

In this section, we first examine the efficacy of the proposed techniques of locally-connected projections and pyramidal schedules.
We then demonstrate that with these techniques our \method{} further enables superior performances on conditional modeling tasks, including class-conditional generation and image inpainting.

We use MNIST, Fashion-MNIST~\cite{xiao2017fashion}, CIFAR-10~\cite{krizhevsky2009learning} and CelebA~\cite{liu2015deep} datasets in our experiments.
For CelebA, we first center-crop the images to $140\times 140$ according to \citet{song2020score} and then resize them to $64\times 64$.
For all experiments, we set $H=10000$ for the number of projections in each step and set the step size $\eta=d$.
The number of simulation steps $K$ varies from $10000$ to $20000$ for different datasets, due to different resolutions and pyramidal schedules.
For MNIST and Fashion-MNIST, we set $M=2.5\times10^5$.
For CIFAR-10 and CelebA, we set $M=7\times10^5$ and $M=4.5\times10^5$, respectively.
Additional experimental details and ablation studies are provided in Appendix~\ref{app:exp-detail}~\&~\ref{app:exp:abla}.
Code is available at 
\href{https://github.com/duchao0726/Conditional-SWF}{https://github.com/duchao0726/Conditional-SWF}.

\subsection{Unconditional Generation}\label{sec:exp:uncond}

\begin{figure}[t]
\centering
\includegraphics[width=\linewidth]{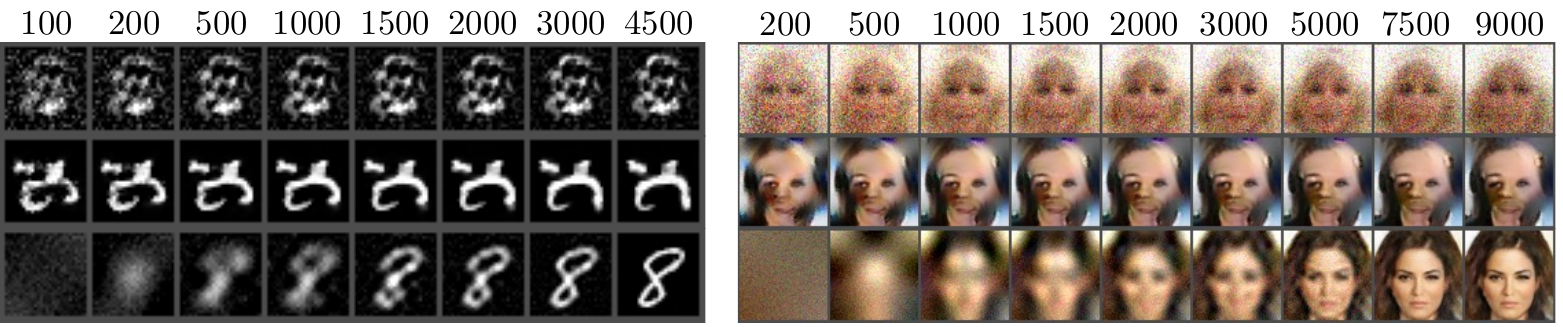}
\caption{Ablation study of the proposed locally-connected projections and pyramidal schedules.
Initially, uniformly sampled projections leads to slow convergence (top rows).
Using locally-connected projections, the samples converge rapidly but lose semantic information (middle rows).
Further combined with the pyramidal schedules, it is possible to generate high-quality samples quickly (bottom rows).
Numbers indicate the simulation steps.}
\label{fig:local_pyramid}
\vspace{-0.1cm}
\end{figure}

\begin{table}[t]
\centering
\small
\caption{FID$\downarrow$ scores obtained by \ucmethod{} on CIFAR-10 and CelebA.
$\diamond$ Use $160\times 160$ center-cropping. $*$ Use $128\times 128$ center-cropping. $\dagger$ Use $140\times 140$ center-cropping.}
\vskip 0.1in
\label{table:fid}
\setlength{\tabcolsep}{3.6pt}
\begin{tabular}{ccc}
    \toprule
    Method & CIFAR-10   & CelebA  \\
    \midrule
    \textit{Auto-encoder based} & & \\
    VAE~\cite{kingma2013auto} & $155.7$ & $85.7^\diamond$     \\
    SWAE~\cite{wu2019sliced} & $107.9$ & $48.9^*$     \\
    WAE~\cite{tolstikhin2017wasserstein} & $-$ & $42^\dagger$     \\
    CWAE~\cite{knop2020cramer} & $120.0$ &  $49.7^\dagger$    \\
    \midrule
    \textit{Autoregressive \& Energy based} & & \\
    PixelCNN~\cite{van2016conditional} & $65.9$ & $-$     \\
    EBM~\cite{du2019implicit} & $37.9$ & $-$     \\
    \midrule
    \textit{Adversarial} & & \\
    WGAN~\cite{arjovsky2017wasserstein} & $55.2$ & $41.3^\diamond$     \\
    WGAN-GP~\cite{gulrajani2017improved} & $55.8$ & $30.0^\diamond$     \\
    CSW~\cite{nguyen2022revisiting} & $36.8$  & $-$ \\
    SWGAN~\cite{wu2019sliced} & $17.0$  & $13.2^*$ \\
    \midrule
    \textit{Score based} & & \\
    NCSN~\cite{song2019generative} &  $25.3$ & $-$ \\
    \midrule
    \textit{Nonparametric} & & \\
    SWF~\cite{liutkus2019sliced} & $>200$ & $>150^\dagger$     \\
    SINF~\cite{dai2021sliced} & $66.5$ & $37.3^*$  \\
    \ucmethod{} (Ours) & $59.7$ & $38.3^\dagger$     \\
    \bottomrule
\end{tabular}
\vspace{-0.1cm}
\end{table}

\begin{figure}[t]
\centering
    \begin{subfigure}[t]{0.23\textwidth}
        \includegraphics[width=\textwidth]{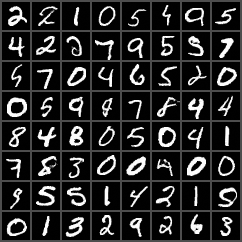} %
    \end{subfigure}
    \begin{subfigure}[t]{0.23\textwidth}
        \includegraphics[width=\textwidth]{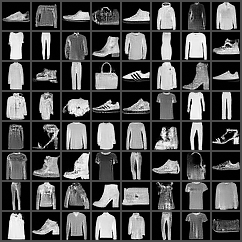} %
    \end{subfigure}
    \begin{subfigure}[t]{0.23\textwidth}
        \includegraphics[width=\textwidth]{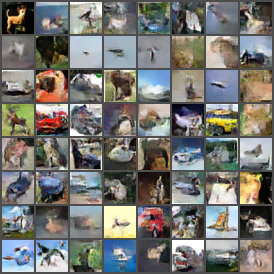} %
    \end{subfigure}
    \begin{subfigure}[t]{0.23\textwidth}
        \includegraphics[width=\textwidth]{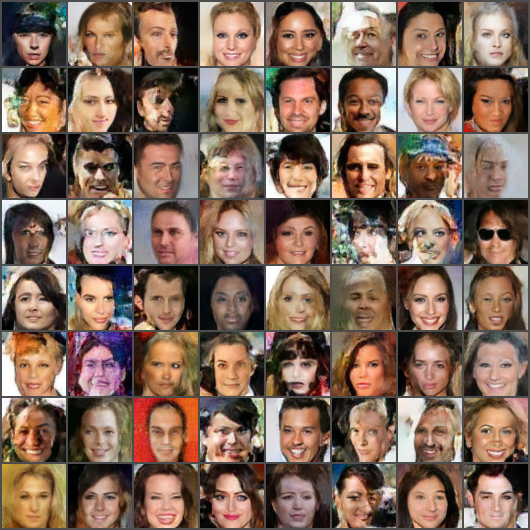} %
    \end{subfigure}
\caption{Uncurated batched samples from \ucmethod{} on MNIST, Fashion MNIST, CIFAR-10 and CelebA.}
\label{fig:uncond-online}
\vspace{-0.1cm}
\end{figure}

To assess the effectiveness of the inductive biases introduced by the locally-connected projections and the pyramidal schedules,
we opt to first evaluate \ucmethod{} on standard unconditional image generation tasks.
We do so because
this makes more existing generative models comparable since most of them are designed for unconditional generation.

\figref{uncond-online} shows uncurated batched samples (see \secref{cswf:algo}) from \ucmethod{} on MNIST, Fashion MNIST, CIFAR-10 and CelebA.
More samples, including their nearest neighbors in the datasets and the offline samples, are shown in Appendix~\ref{app:exp:sample}.
We observe that the generated images are of high quality.
To intuit the effectiveness of locally-connected projections and pyramidal schedules, we show in \figref{local_pyramid} an ablation study.
It can be observed that with the introduced inductive biases,
the number of simulation steps can be greatly reduced, and the generative quality is significantly improved.
For comparison, \citet{liutkus2019sliced} report that uniformly sampled projections (i.e.\ without the inductive biases) fail to produce satisfactory samples on high-dimensional image data.

\begin{figure*}[t]
\centering
    \begin{subfigure}[t]{0.30\textwidth}
        \includegraphics[width=\textwidth]{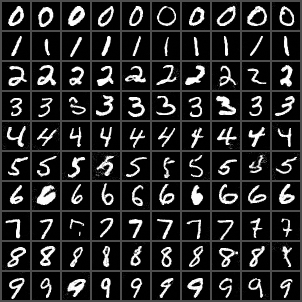} %
    \end{subfigure}
    \begin{subfigure}[t]{0.30\textwidth}
        \includegraphics[width=\textwidth]{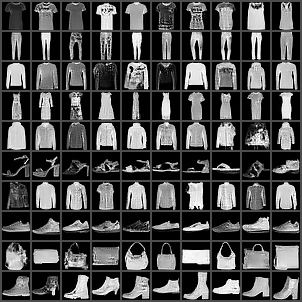} %
    \end{subfigure}
    \begin{subfigure}[t]{0.30\textwidth}
        \includegraphics[width=\textwidth]{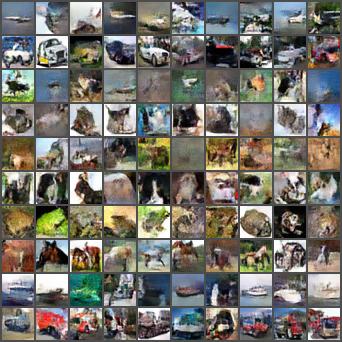} %
    \end{subfigure}
\caption{Class-conditional samples from \method{} ($\xi=10$) on MNIST, Fashion MNIST and CIFAR-10.}
\label{fig:cond-online}
\vspace{-0.1cm}
\end{figure*}

\begin{figure*}[t]
\centering
\includegraphics[width=\textwidth]{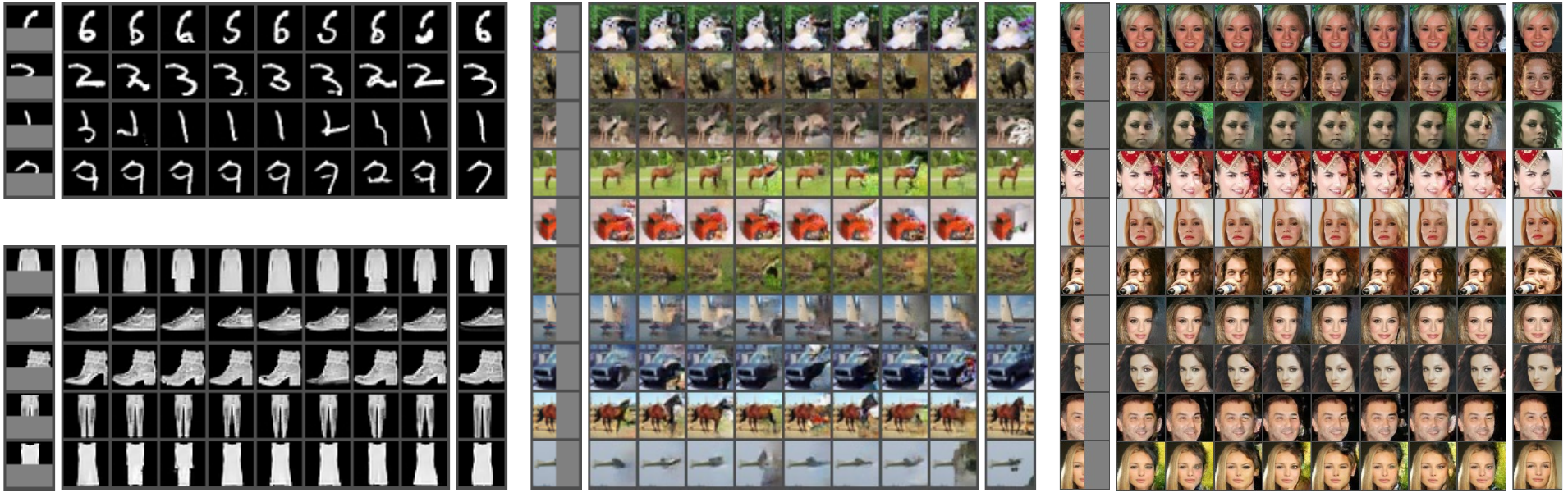} %
\caption{Image inpainting results of \method{} ($\xi=1$) on MNIST, Fashion MNIST, CIFAR-10 and CelebA. In each figure, the leftmost column shows the occluded images and the rightmost column shows the original images.}
\label{fig:inp}
\end{figure*}

We report the FID scores~\cite{heusel2017gans} on CIFAR-10 and CelebA in \tabref{fid} for quantitative evaluation.
We compare with SWF~\cite{liutkus2019sliced} and SINF~\cite{dai2021sliced}, which are also iterative methods based on the $SW_2$ distance.
We also include the results of other generative models based on optimal transport for comparison, including SWAE \& SWGAN~\cite{wu2019sliced}, WAE~\cite{tolstikhin2017wasserstein}, CWAE~\cite{knop2020cramer}, CSW~\cite{nguyen2022revisiting}, WGAN~\cite{arjovsky2017wasserstein} and WGAN-GP~\cite{gulrajani2017improved}.
For better positioning our method in the literature of generative models,
we also list results of some representative works, including VAE~\cite{kingma2013auto}, PixelCNN~\cite{van2016conditional}, EBM~\cite{du2019implicit} and NCSN~\cite{song2019generative}.
Our \ucmethod{} significantly outperforms SWF due to the appropriately introduced inductive biases via the techniques described in \secref{convcswf}.
On CIFAR-10, it also outperforms SINF (which is layer-wise optimized),
probably because SINF requires the projections to be orthogonal, which limits its capability.
We include results on CelebA for reference, while we note that different preprocessing make the scores not directly comparable.
It is worth noting that, as a nonparametric method that does not require any optimization (e.g.\ backpropagation), \ucmethod{} achieves comparable results to many elaborate parametric methods such as WGAN and PixelCNN, showing great promise.

\subsection{Conditional Modeling}\label{sec:exp:cond}

We now demonstrate that, with the help of the introduced inductive biases, our \method{} is capable of handling commonly concerned conditional distribution modeling tasks such as class-conditional generation and image inpainting.

\subsubsection{Class-Conditional Image Generation}

For class-conditional generation tasks, we let $\mathcal{Y}$ be the set of one-hot vectors representing the class labels.
The initial $\{\bar{Y}_0^j\}_{j=1}^M$ are sampled according to the label distribution in the dataset (which is categorically uniform for all three datasets tested here).
For each projection $\theta_x$, we additionally sample a $\theta_y$ uniformly from $\mathbb{S}^{l-1}$ with $l$ being the number of classes and then normalize it together with $\theta_x$, ensuring that $\theta=[\theta_x^\top,\theta_y^\top]^\top$ has unit length.
We set the amplifier $\xi=10$ for all datasets.
Other experimental settings, including the hyperparameters and the pyramidal schedules, are the same as in \secref{exp:uncond}.
The generated images are shown in \figref{cond-online}.
We observe that the samples are of good visual quality and consistent with the class labels.
Interestingly, by varying the amplifier $\xi$, \method{} can smoothly transit between class-conditional and unconditional generation, as shown in Appendix~\ref{app:exp:abla}.

\subsubsection{Image Inpainting}

For inpainting tasks, we let $\mathcal{X}$ and $\mathcal{Y}$ represent the pixel spaces of the occluded and observed portions of images, respectively.
Since the true marginal $q(y)$ is not available in this setting,
we set the initial $\{\bar{Y}_0^j\}_{j=1}^M$ to the partially-observed images created from the dataset.
We directly sample $\theta$ (using locally-connected projections) rather than dealing with $\theta_x$ and $\theta_y$ separately, as both $\mathcal{X}$ and $\mathcal{Y}$ are in the image domain.
The amplifier is set to $\xi=1$ for all datasets.
In \figref{inp}, we show inpainting results (offline samples) for the occluded images created from the test split of each dataset.
We observe that the inpaintings are semantically meaningful and consistent with the given pixels.

\section{Conclusions}

In this work, we make two major improvements to SWF, a promising type of nonparametric generative model. We first extend SWF in a natural way to support conditional distribution modeling, which opens up the possibility to applications that rely on conditional generation, e.g.\ text-to-image, image inpainting. On the other hand, we introduce domain-specific inductive biases for image generation, which significantly improves the efficiency and generative quality. Despite being a nonparametric model that does not rely on backpropagation training, our method performs comparably to many parametric models. This promising performance could inspire more further research in this field.

\bibliography{ref}
\bibliographystyle{icml2023}

\newpage
\appendix
\onecolumn

\begin{figure}
    \centering
    \begin{subfigure}[t]{0.99\textwidth}
        \includegraphics[width=\textwidth]{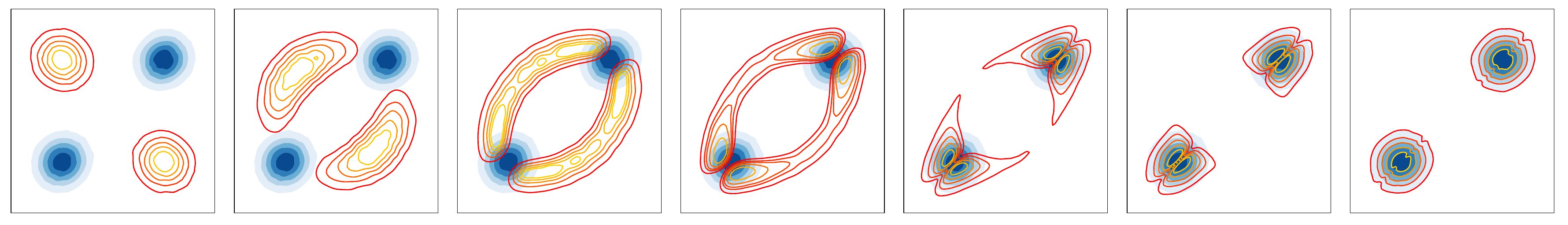}
        \caption{The joint SWF.}\label{fig:illustration:swf1}
    \end{subfigure}
    \begin{subfigure}[t]{0.99\textwidth}
        \includegraphics[width=\textwidth]{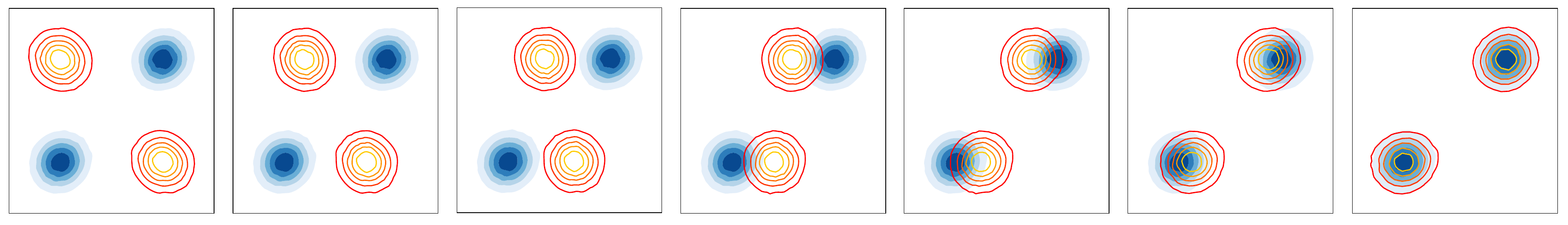}
        \caption{The ideal conditional SWFs.}\label{fig:illustration:idea_cswf}
    \end{subfigure}
    \begin{subfigure}[t]{0.99\textwidth}
        \includegraphics[width=\textwidth]{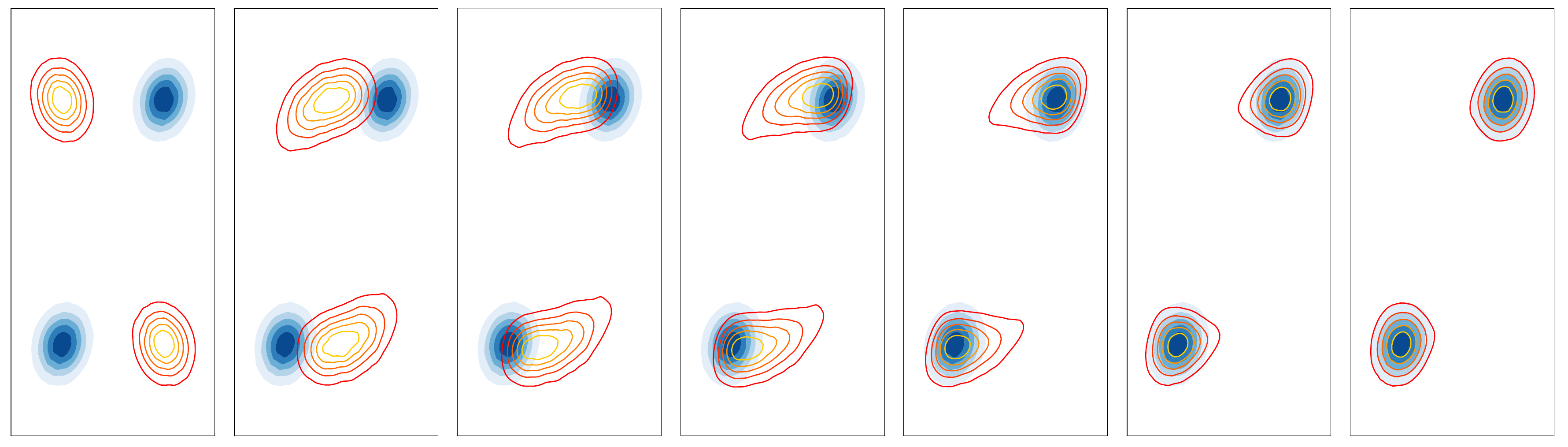}
        \caption{The joint SWF after moving components farther apart in the $y$-direction.}\label{fig:illustration:swf2}
    \end{subfigure}
    \begin{subfigure}[t]{0.99\textwidth}
        \includegraphics[width=\textwidth]{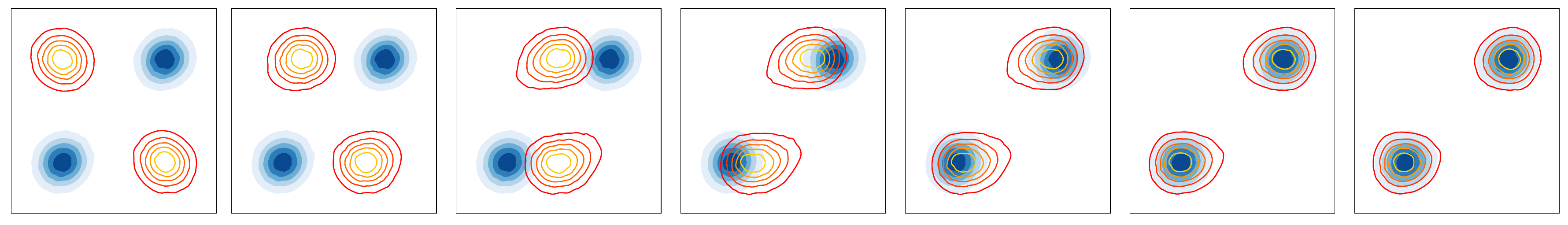}
        \caption{Illustration of the proposed CSWF.}\label{fig:illustration:cswf}
    \end{subfigure}
    \begin{subfigure}[t]{0.99\textwidth}
        \includegraphics[width=\textwidth]{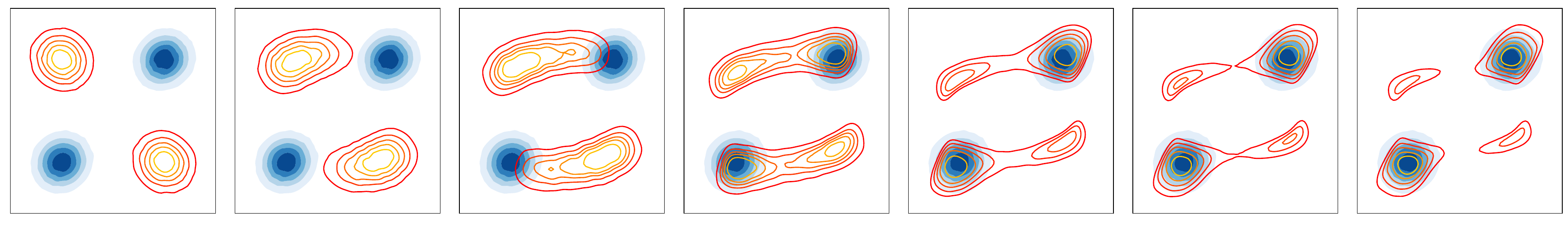}
        \caption{Illustration of the proposed CSWF without amplifying (i.e., with $\xi=1$).}\label{fig:illustration:cswf_noxi}
    \end{subfigure}
    \caption{Illustrations of joint SWFs, conditional SWFs and the proposed CSWF algorithm. See more explanations in Appendix~\ref{app:illustration}.}
    \label{fig:illustration}
\end{figure}

\section{Illustrations of Joint SWFs, Conditional SWFs and CSWF}\label{app:illustration}

In this section, we show several 2-dimensional toy examples (i.e.\ $\mathcal{X}=\mathcal{Y}=\R$) to motivate our CSWF method.

\textbf{The Joint SWF} Suppose we have a target distribution $q(x,y)\in\mathcal{P}_2(\mathbb{R}^2)$, which is a mixture of two Gaussian distributions (outlined in blue shaded contours in \figref{illustration:swf1}) and an initial distribution $p_0(x,y)\in\mathcal{P}_2(\mathbb{R}^2)$, which is a mixture of another two Gaussian distributions with different modes (outlined in red contours in \figref{illustration:swf1}).
The joint SWF $(p_t)_{t\geq0}$ starting from $p_0$ and targeting $q$ is demonstrated in \figref{illustration:swf1}, chronologically from left to right. We observe that each mixture component of $p_0$ is ``split'' into two parts and moves to two different components of the target distribution $q$.

\textbf{The Ideal Conditional SWFs} In the setting of conditional modeling described in \secref{cswf}, we aim to fit $q(x|y)$ for all $y\in\mathcal{Y}$.
Ideally, we can achieve this with a conditional SWF starting from $p_0(x|y)$ and targeting $q(x|y)$ for each $y$, as described in \secref{cswf:naivecswf}.
\figref{illustration:idea_cswf} illustrates the effect of the ideal conditional SWFs $(p_{y,t}(x))_{t\geq0},\forall y\in\mathcal{Y}$.

\textbf{A Difference Joint SWF} We now alter the initial distribution $p_0(x,y)$ and the target $q(x,y)$ by simply shifting their mixture components farther apart in the $y$-direction (i.e.,\ the vertical direction), and then show the new joint SWF in \figref{illustration:swf2}.
The mixture components now move (roughly) horizontally, resulting in a significantly different trajectory than in \figref{illustration:swf1}.

\textbf{Motivation of CSWF} While the above example (\figref{illustration:swf2}) remains a joint SWF, it bears considerable resemblance to the desired conditional SWFs (\figref{illustration:idea_cswf}). This motivates us to approximate conditional SWFs with a joint SWF of scaled initial and target distributions (Algorithm~\ref{algo:cswf}). Specifically, we first stretch the initial and target distributions along the $\mathcal{Y}$-component (using a factor $\xi$, which we call an amplifier), then simulate the joint SWF of the streched initial and target distributions, and finally compress the distributions to the original scale.
We show the results of our CSWF in \figref{illustration:cswf}.

\textbf{Significance of $\xi$ in CSWF}
Note that the effect of a large amplifier $\xi$ is significant, since scaling the $\mathcal{Y}$-component is the key factor in making the joint SWF approximate the conditional SWFs.
In \figref{illustration:cswf_noxi}, we show the results of CSWF without amplifying (i.e., with $\xi=1$) for comparison.

\section{Recap of the SWF Algorithm~\cite{liutkus2019sliced}}\label{app:swf}

\citet{liutkus2019sliced} consider minimizing the functional $\F_\lambda^q(\cdot)=\frac{1}{2}SW^2_2(\cdot,q) + \lambda\mathcal{H}(\cdot)$, where $\mathcal{H}(\cdot)$ denotes the negative entropy defined by $\mathcal{H}(p)\triangleq\int_{\Rd}p(x)\log p(x)\dd x$. The introduced entropic regularization term helps to have the convergence of the Wasserstein gradient flow. In specific, they prove that under certain conditions the Wasserstein gradient flow of $\F_\lambda^q$ admits density $(p_t)_{t\geq0}$ that satisfies the following continuity equation:
\begin{equation*}
\frac{\partial p_t(x)}{\partial t} + \nabla\cdot(p_t(x)v_t(x)) - \lambda\Delta p_t=0,\quad v_t(x)\triangleq-\int_{\Sd}\psi'_{t,\theta}(\theta^\top x)\cdot\theta~\dd\lambda(\theta).
\end{equation*}
Compared to \eqn{swfpde}, there is an extra Laplacian term which corresponds to the entropic regularization in $\F_\lambda^q$.
By drawing a connection between this Fokker-Planck-type equation and stochastic differential equations (SDE), they propose to simulate the above equation with a stochastic process $\dd X_t=v_t(X_t)\dd t + \sqrt{2\lambda}\dd W_t$, where $(W_t)_t$ denotes a standard Wiener process.
Finally, they propose to approximate the SDE with a particle system and present a practical algorithm for unconditional generative modeling, which we recap in Algorithm~\ref{algo:swf} (using our notations).

\begin{algorithm2e}[tb]
    \SetInd{0.1ex}{1.5ex}
    \DontPrintSemicolon
    \SetKwInput{Input}{Input}
    \SetKwInput{Output}{Output}
    \Input{$\mathcal{D}=\{x_i\}_{i=1}^N$, $M$, $H$, $\eta$, $\lambda$, $K$}
    \Output{$\{\bar{X}_K^j\}_{j=1}^M$}
    {\color{blue} \small \tcp{Initialize the particles}}
    $\{\bar{X}_0^j\}_{j=1}^M \simiid \mathcal{N}(0,I_d)$\\
    {\color{blue} \small \tcp{Generate random directions}}
    $\{\theta_{h}\}_{h=1}^H \simiid \mathrm{Uniform}(\Sd)$\\
    {\color{blue} \small \tcp{Quantiles of projected target}}
    \For{$h = 1,\dots,H$}
    {
        $F_{\theta_{h\sharp}^* \hat{q}}^{-1}=\textrm{QF}(\{\theta_{h}^\top x_i\}_{i=1}^N)$ ~~~~{\color{blue} \small \tcp{QF denotes the quantile function}}
    }
    {\color{blue} \small \tcp{Iterations}}
    \For{$k = 0,\dots, K-1$}
    {
        \For{$h = 1,\dots,H$}
        {
        {\color{blue} \small \tcp{CDF of projected particles}}
        $F_{\theta_{h\sharp}^* \hat{p}_k}=\textrm{CDF}(\{\theta_{h}^\top \bar{X}_k^j\}_{j=1}^M)$\\
        }
        \For{$j=1,\dots,M$}
        {
        {\color{blue} \small \tcp{Update the particles}}
        $\bar{X}_{k+1}^j = \bar{X}_{k}^j - \eta \cdot\hat{v}_k(\bar{X}_{k}^j) + \sqrt{2\lambda\eta}\cdot\epsilon_{k+1}^j,\quad\quad\epsilon_{k+1}^j\sim\mathcal{N}(0,I_d)$
        }
    }
    \caption{Sliced-Wasserstein Flow (SWF)~\cite{liutkus2019sliced}}
    \label{algo:swf}
\end{algorithm2e}

\section{Additional Experimental Details}\label{app:exp-detail}

In our experiments, 
we augment the CIFAR-10 dataset with horizontally flipped images, resulting in a total of $100000$ training images. This is analogous to the random flip data augmentation used in training neural networks.
We do not employ this augmentation for the CelebA dataset due to limited computing resources.
The pixel values of all images are dynamically dequantized at each step during the simulation and are rescaled to the range of $[-1,1]$.

We set the simulation step size $\eta=d$ (i.e.\ the dimensionality of the images) due to the following reasons.
In one-dimensional cases ($d=1$), we can solve the optimal transport problem between $\thetapush p_t$ and $\thetapush q$ using \eqn{v-empirical} with step size $\eta=1$, since it recovers the optimal transport map $T$.
In $d$-dimensional cases, for any $d$ orthogonal projections, the optimal transport problems between the projected distributions are independent of each other and can be solved simultaneously.
In \eqn{v-empirical}, however, the transport maps (i.e., the derivatives of the Kantorovich potentials) of all directions are averaged.
Therefore, we set $\eta=d$ so that the step size in each direction equals to $1$ in the average sense.

\begin{table}[t]
\centering
\caption{Details of the pyramidal schedules for each dataset. In each entry, $(\mathtt{H}\times\mathtt{W})~[\mathtt{S}_1,\dots,\mathtt{S}_k]$ denotes that we use locally-connected projections of resolution $\mathtt{H}\times\mathtt{W}$ with patch size $\mathtt{S}_1\times\mathtt{S}_1,\dots,\mathtt{S}_k\times \mathtt{S}_k$ sequentially. We upsample all projections to image resolution.}
\vskip 0.1in
\label{table:layers}
\setlength{\tabcolsep}{8pt}
\begin{tabular}{lll}
    \toprule
    MNIST \& Fashion MNIST &  CIFAR-10   & CelebA  \\
    \midrule
    $(1\times 1)~[1]$ & $(1\times 1)~[1]$ & $(1\times 1)~[1]$ \\
    $(2\times 2)~[2]$ & $(2\times 2)~[2]$ & $(2\times 2)~[2]$ \\
    $(3\times 3)~[3]$ & $(3\times 3)~[3]$ & $(3\times 3)~[3]$ \\
    $(4\times 4)~[4]$ & $(4\times 4)~[4]$ & $(4\times 4)~[4]$ \\
    $(5\times 5)~[5]$ & $(5\times 5)~[5]$ & $(5\times 5)~[5]$ \\
    $(6\times 6)~[6]$ & $(6\times 6)~[6]$ & $(6\times 6)~[6]$ \\
    $(7\times 7)~[7,5,3]$ & $(7\times 7)~[7]$ & $(7\times 7)~[7]$ \\
    $(11\times 11)~[11,9,7,5,3]$ & $(8\times 8)~[8,7,5,3]$ & $(8\times 8)~[8,7,5,3]$ \\
    $(14\times 14)~[14,13,11,9,7,5,3]$ & $(12\times 12)~[12,11,9,7,5,3]$ & $(12\times 12)~[12,11,9,7,5,3]$ \\
    $(21\times 21)~[15,13,11,9,7,5,3]$ & $(16\times 16)~[15,13,11,9,7,5,3]$ & $(16\times 16)~[15,13,11,9,7,5,3]$ \\
    $(28\times 28)~[15,13,11,9,7,5,3]$ & $(24\times 24)~[15,13,11,9,7,5,3]$ & $(24\times 24)~[15,13,11,9,7,5,3]$ \\
     & $(32\times 32)~[15,13,11,9,7,5,3]$ & $(32\times 32)~[15,13,11,9,7,5,3]$ \\
     &  & $(64\times 64)~[15,13,11,9,7,5,3]$ \\
    \bottomrule
\end{tabular}
\end{table}

\subsection{Locally-Connected Projections and Pyramidal Schedules}

We summarize the pyramidal schedules used for each dataset in \tabref{layers}.
When upsampling projections with a lower resolution than the image,
we empirically find that Lanczos upsampling works better than nearest neighbor upsampling.

\subsection{CDF Estimations}

We estimate the CDFs of the projected distributions $\thetapush \hat{p}_t$ and $\thetapush \hat{q}$ by first sorting their (scalar) projected values $\{\theta_{x}^{\top} \bar{X}^j+\xi\cdot\theta_{y}^{\top} \bar{Y}^j \}_{j=1}^M$ and $\{\theta_{x}^{\top} x_i+\xi\cdot\theta_{y}^{\top} y_i \}_{i=1}^N$, respectively. After sorting, the linear interpolation is performed as follows.
Let $\{z_i\}_{i=1}^N$ denote the sorted array, i.e., $z_1\leq\cdots\leq z_N$. When estimating the CDF of an input value $z'$, we first find its insert position $I$ (i.e., the index $I$ satisfying $z_I\leq z'\leq z_{I+1}$) with binary search. Then the CDF of $z'$ is estimated with $\frac{I-1}{N}+\frac{1}{N}\frac{z'-z_I}{z_{I+1}-z_I}$.
For an input $a\in[0,1]$, we inverse the CDF by first calculating the index $I=\floor*{a \times N}$ and then computing the inverse value as $z_I+(a\times N-I)*(z_{I+1}-z_{I})$.
For CIFAR-10 and CelebA, since we use a relatively large number of particles which leads to a slow sorting procedure, we choose a subset of particles for the estimation of $\thetapush \hat{p}_t$.

\section{Additional Experiments}\label{app:exp}

\subsection{Ablation Studies of $H$, $M$ and $\xi$}\label{app:exp:abla}

\begin{table}[t]
\centering
\caption{FID$\downarrow$ scores of \ucmethod{} using different number of Monte Carlo samples $H$ on CIFAR-10.}
\vskip 0.1in
\label{table:abla-mcsamples}
\setlength{\tabcolsep}{8pt}
\begin{tabular}{cc}
    \toprule
    \# Monte Carlo Samples  &  FID   \\
    \midrule
    $H = 1000$ & $90.8$\\
    $H = 5000$ & $68.1$\\
    $H = 10000$ & $59.7$ \\
    \bottomrule
\end{tabular}
\end{table}

\begin{table}[t]
\centering
\caption{FID$\downarrow$ scores of \ucmethod{} using different number of particles $M$ on CIFAR-10.}
\vskip 0.1in
\label{table:abla-particles}
\setlength{\tabcolsep}{8pt}
\begin{tabular}{cc}
    \toprule
    \# Particles  &  FID \\
    \midrule
    $M = 1\times 10^5$ & $64.0$ \\
    $M = 3\times 10^5$ & $61.4$ \\
    $M = 7\times 10^5$ & $59.7$ \\
    \bottomrule
\end{tabular}
\end{table}

\begin{figure}[t]
\centering
    \begin{subfigure}[t]{0.28\textwidth}
        \includegraphics[width=\textwidth]{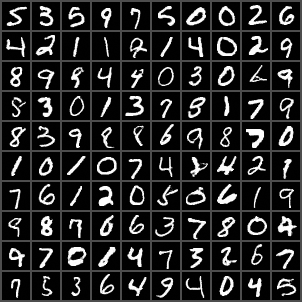}
        \caption{MNIST, $\xi=0$}
    \end{subfigure}
    \begin{subfigure}[t]{0.28\textwidth}
        \includegraphics[width=\textwidth]{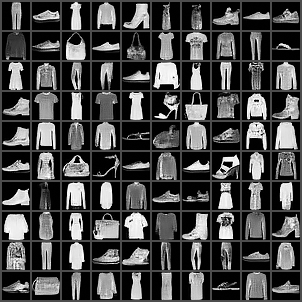}
        \caption{Fashion MNIST, $\xi=0$}
    \end{subfigure}
    \begin{subfigure}[t]{0.28\textwidth}
        \includegraphics[width=\textwidth]{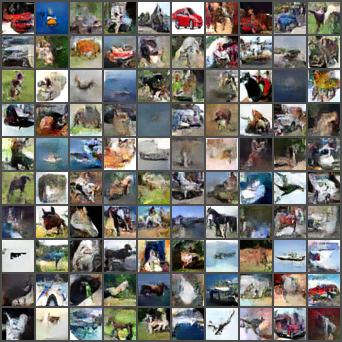}
        \caption{CIFAR-10, $\xi=0$}
    \end{subfigure}
    \begin{subfigure}[t]{0.28\textwidth}
        \includegraphics[width=\textwidth]{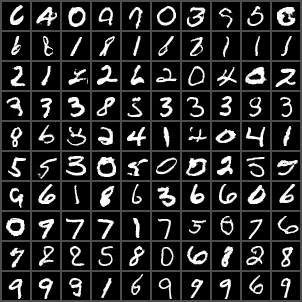}
        \caption{MNIST, $\xi=2$}
    \end{subfigure}
    \begin{subfigure}[t]{0.28\textwidth}
        \includegraphics[width=\textwidth]{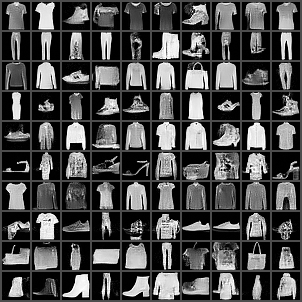}
        \caption{Fashion MNIST, $\xi=2$}
    \end{subfigure}
    \begin{subfigure}[t]{0.28\textwidth}
        \includegraphics[width=\textwidth]{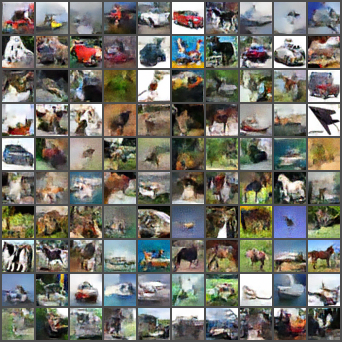}
        \caption{CIFAR-10, $\xi=2$}
    \end{subfigure}
    \begin{subfigure}[t]{0.28\textwidth}
        \includegraphics[width=\textwidth]{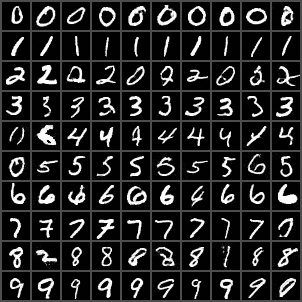}
        \caption{MNIST, $\xi=5$}
    \end{subfigure}
    \begin{subfigure}[t]{0.28\textwidth}
        \includegraphics[width=\textwidth]{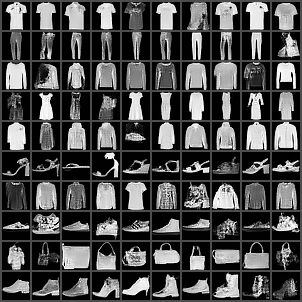}
        \caption{Fashion MNIST, $\xi=5$}
    \end{subfigure}
    \begin{subfigure}[t]{0.28\textwidth}
        \includegraphics[width=\textwidth]{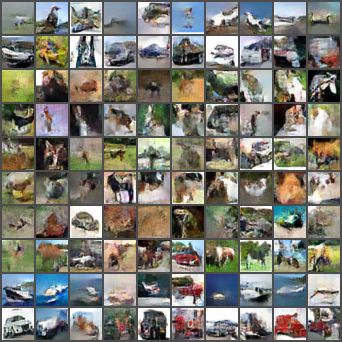}
        \caption{CIFAR-10, $\xi=5$}
    \end{subfigure}
    \begin{subfigure}[t]{0.28\textwidth}
        \includegraphics[width=\textwidth]{mnist_cond_online_xi10.png}
        \caption{MNIST, $\xi=10$}
    \end{subfigure}
    \begin{subfigure}[t]{0.28\textwidth}
        \includegraphics[width=\textwidth]{fashion_cond_online_xi10.png}
        \caption{Fashion MNIST, $\xi=10$}
    \end{subfigure}
    \begin{subfigure}[t]{0.28\textwidth}
        \includegraphics[width=\textwidth]{cifar_cond_online_xi10.png}
        \caption{CIFAR-10, $\xi=10$}
    \end{subfigure}
  \vspace{-.2cm}
\caption{Class-conditional generation of \method{} with different amplifiers $\xi$ on MNIST, Fashion MNIST and CIFAR-10. In each figure, each row corresponds to a class. We observe that $\xi=0$ recovers unconditional generation and the generated samples become more consistent with the classes as $\xi$ grows.}
\label{fig:amplifier-ablation}
\end{figure}

We present the FID scores obtained by \ucmethod{} using different numbers of Monte Carlo samples $H$ in \tabref{abla-mcsamples}. The results of \ucmethod{} using different numbers of particles $M$ are shown in \tabref{abla-particles}.
We show the class-conditional generation of \method{} using different amplifiers $\xi$ from $0$ to $10$ in \figref{amplifier-ablation}.

\subsection{Additional Samples}\label{app:exp:sample}

\begin{figure}[t]
\centering
    \begin{subfigure}[t]{0.45\textwidth}
        \includegraphics[width=\textwidth]{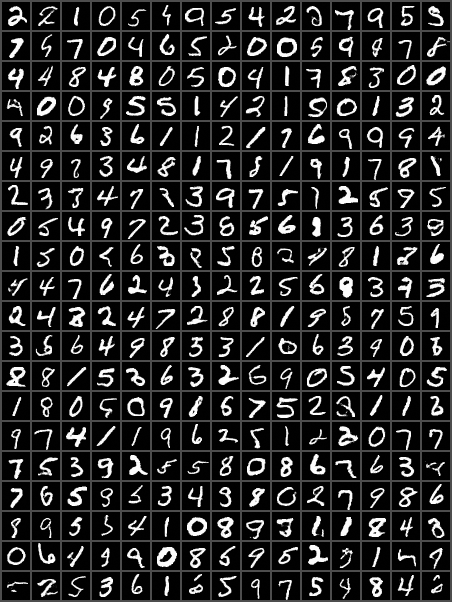} %
        \caption{MNIST}
    \end{subfigure}
    \begin{subfigure}[t]{0.45\textwidth}
        \includegraphics[width=\textwidth]{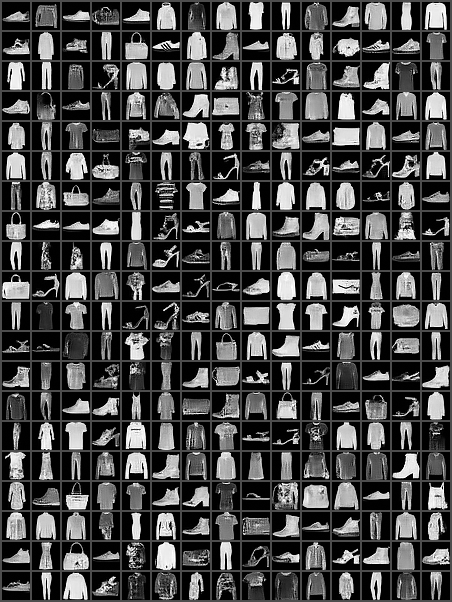} %
        \caption{Fashion MNIST}
    \end{subfigure}
    \begin{subfigure}[t]{0.45\textwidth}
        \includegraphics[width=\textwidth]{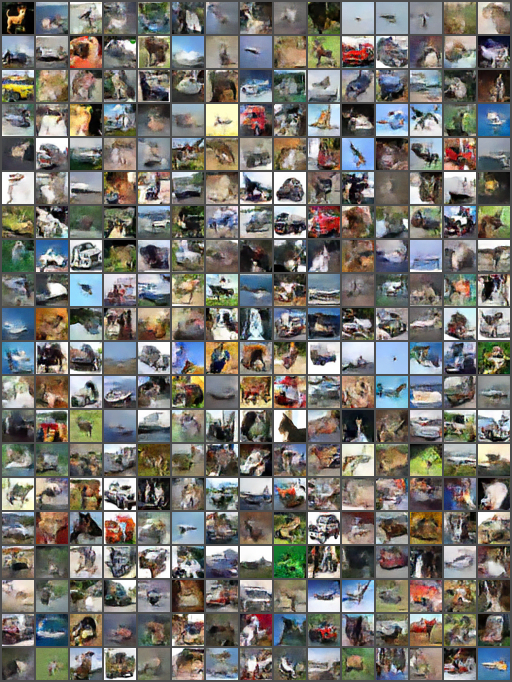} %
        \caption{CIFAR-10}
    \end{subfigure}
    \begin{subfigure}[t]{0.45\textwidth}
        \includegraphics[width=\textwidth]{celeba_uncond_online_more_compress.pdf} %
        \caption{CelebA}
    \end{subfigure}
\caption{Additional uncurated batched samples from \ucmethod{} on MNIST, Fashion MNIST, CIFAR-10 and CelebA.}
\label{fig:uncond-online-more}
\end{figure}

\begin{figure}[t]
\centering
\begin{subfigure}[t]{0.35\textwidth}
    \includegraphics[width=\textwidth]{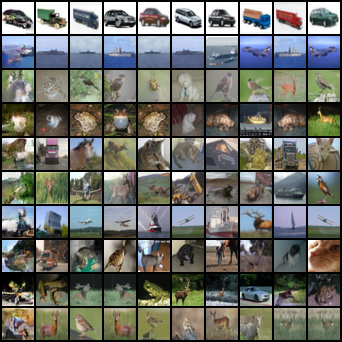} %
    \caption{CIFAR-10}
\end{subfigure}
\hspace{.5cm}
\begin{subfigure}[t]{0.35\textwidth}
    \includegraphics[width=\textwidth]{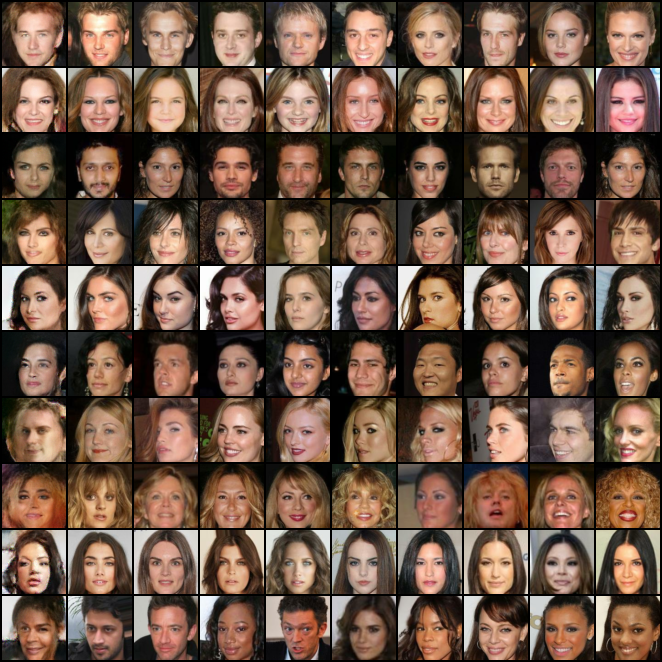} %
    \caption{CelebA}
\end{subfigure}
\caption{$L_2$ nearest neighbors of the generated samples from \ucmethod{} on CIFAR-10 and CelebA.
The leftmost columns are the generated images. Images to the right are the nearest neighbors in the dataset.}
\label{fig:uncond-online-nn}
\end{figure}

\begin{figure}[t]
\centering
    \begin{subfigure}[t]{0.23\textwidth}
        \includegraphics[width=\textwidth]{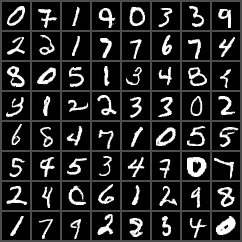} %
        \caption{MNIST}
    \end{subfigure}
    \begin{subfigure}[t]{0.23\textwidth}
        \includegraphics[width=\textwidth]{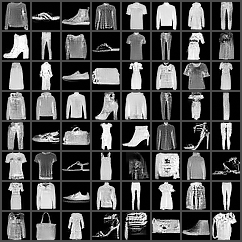} %
        \caption{Fashion MNIST}
    \end{subfigure}
    \begin{subfigure}[t]{0.23\textwidth}
        \includegraphics[width=\textwidth]{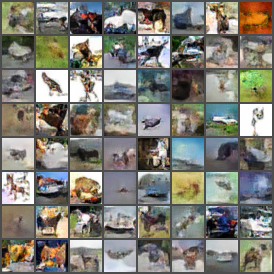} %
        \caption{CIFAR-10}
    \end{subfigure}
    \begin{subfigure}[t]{0.23\textwidth}
        \includegraphics[width=\textwidth]{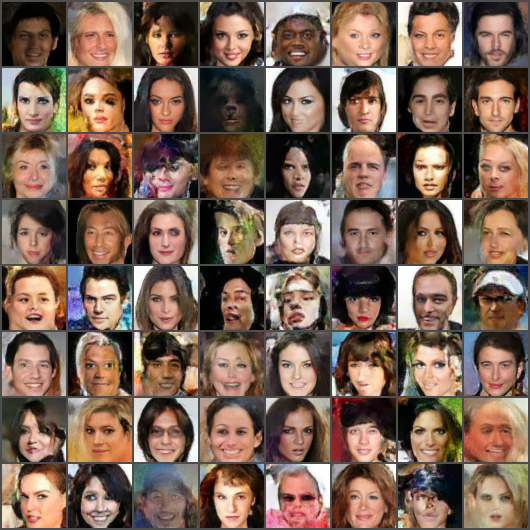} %
        \caption{CelebA}
    \end{subfigure}
\caption{Uncurated offline samples from \ucmethod{} on MNIST, Fashion MNIST, CIFAR-10 and CelebA.}
\label{fig:uncond-offline}
\end{figure}

More unconditional samples from \ucmethod{} are shown in \figref{uncond-online-more}.
We show the nearest neighbors of the generated samples in \figref{uncond-online-nn},
where we observe that the generated samples are not replicated training samples or combined training patches, but generalize at the semantic level.
In \figref{uncond-offline}, we show the offline samples, which appear to be comparable in visual quality to the batch samples (in \figref{uncond-online}).
Quantitatively, the FID score of the offline samples on CIFAR-10 is $61.1$, which is also close to that of the batched samples ($59.7$).

\begin{figure}[t]
\centering
    \begin{subfigure}[t]{0.80\textwidth}
        \includegraphics[width=\textwidth]{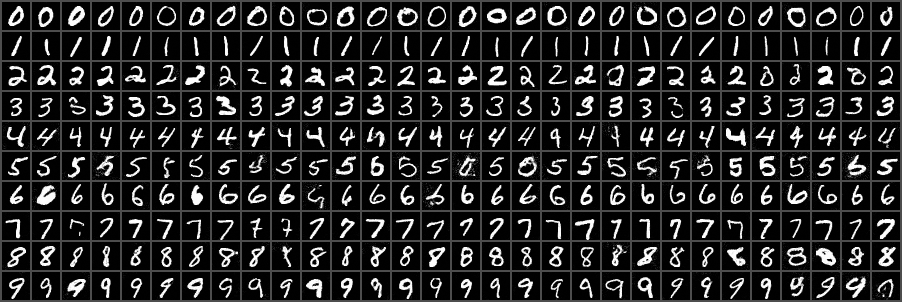} %
        \caption{MNIST}
    \end{subfigure}
    \begin{subfigure}[t]{0.80\textwidth}
        \includegraphics[width=\textwidth]{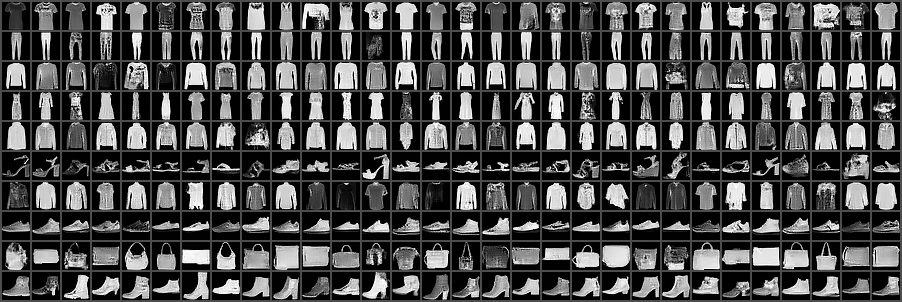} %
        \caption{Fashion MNIST}
    \end{subfigure}
    \begin{subfigure}[t]{0.80\textwidth}
        \includegraphics[width=\textwidth]{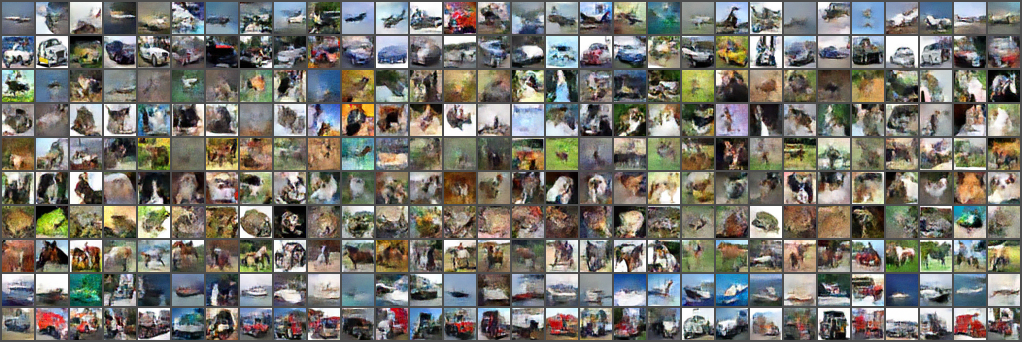} %
        \caption{CIFAR-10}
    \end{subfigure}
\caption{Additional class-conditional samples from \method{} ($\xi=10$) on MNIST, Fashion MNIST and CIFAR-10.}
\label{fig:cond-online-more}
\end{figure}

\begin{figure}[t]
\centering
    \begin{subfigure}[t]{0.68\textwidth}
        \includegraphics[width=\textwidth]{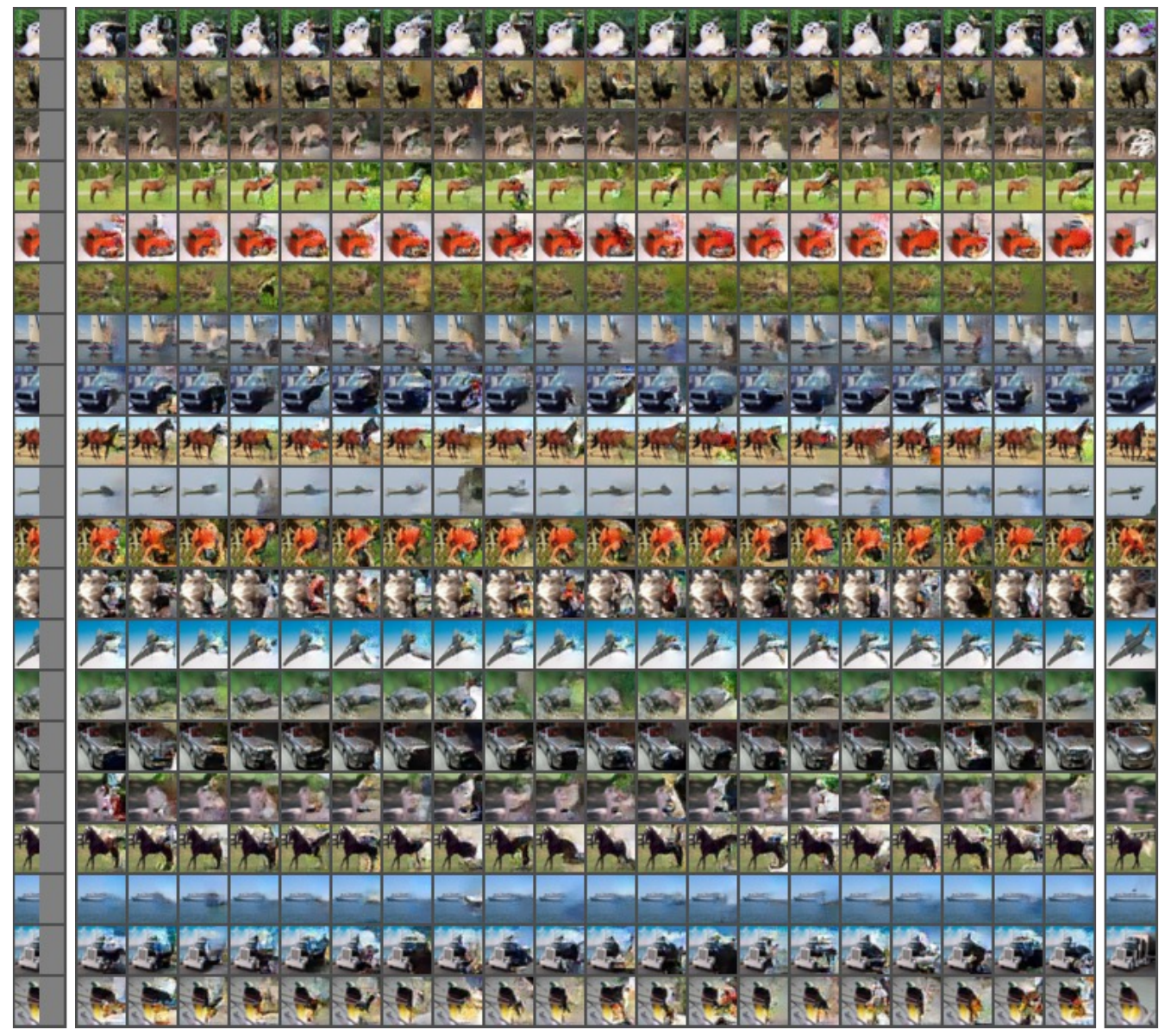} %
        \caption{CIFAR-10}
    \end{subfigure}
    \begin{subfigure}[t]{0.68\textwidth}
        \includegraphics[width=\textwidth]{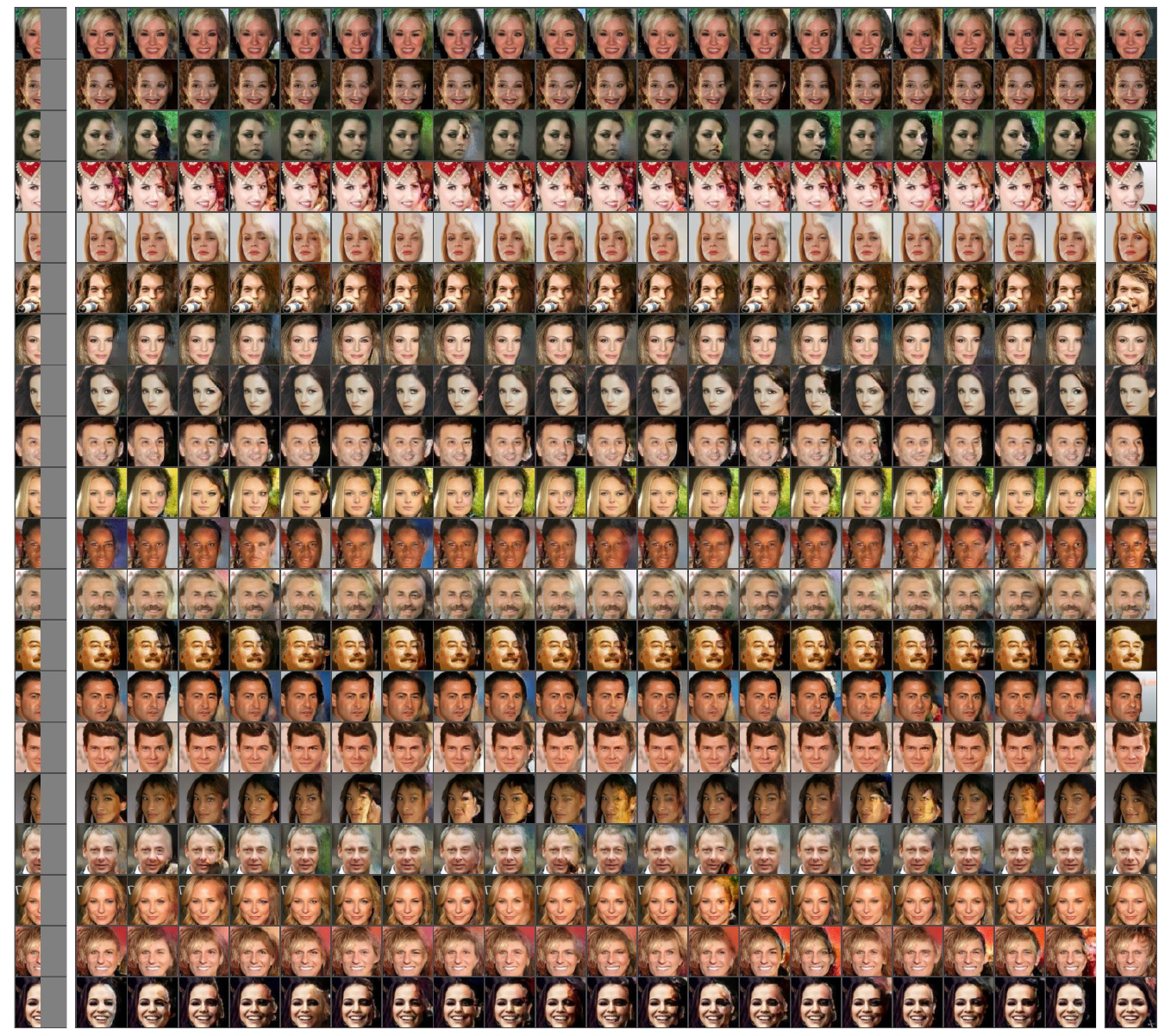} %
        \caption{CelebA}
    \end{subfigure}
\caption{Additional image inpainting results of \method{} ($\xi=1$) on CIFAR-10 and CelebA.}
\label{fig:inp-more}
\end{figure}

\figref{cond-online-more} and \figref{inp-more} show additional class-conditional samples and image inpainting results of \method{}, respectively.

\end{document}